\documentclass[10pt,twocolumn,letterpaper]{article}

\usepackage{cvpr}
\usepackage{times}
\usepackage{epsfig}
\usepackage{graphicx}
\usepackage{amsmath}
\usepackage{amssymb}
\usepackage{booktabs}

\usepackage[linesnumbered,ruled]{algorithm2e}
\usepackage[pagebackref=true,breaklinks=true,letterpaper=true,colorlinks,bookmarks=false]{hyperref}
\usepackage[dvipsnames,table,xcdraw]{xcolor}
\usepackage{arydshln}
\usepackage{float}
\restylefloat{table}
\hypersetup{final}
\usepackage[font=small,labelfont=bf,tableposition=top, skip=0pt]{caption}

\definecolor{ora}{rgb}{1,0.5,0}

\cvprfinalcopy %

\def\cvprPaperID{6188} %

\ifcvprfinal\fi
\begin{document}

\title{RePr: Improved Training of Convolutional Filters}

\author{Aaditya Prakash\thanks{Part of the research was done while the author was an intern at MSR}\\
Brandeis University\\
{\tt\small aprakash@brandeis.edu}
\and
James Storer\\
Brandeis University\\
{\tt\small storer@brandeis.edu}
\and
Dinei Florencio, Cha Zhang\\
Microsoft Research\\
{\tt\small \{dinei,chazhang\}@microsoft.com}
}

\maketitle

\begin{abstract}
A well-trained Convolutional Neural Network can easily be pruned without significant loss of performance. 
This is because of unnecessary overlap in the features captured by the network's filters. 
Innovations in network architecture such as skip/dense connections and Inception units have mitigated this problem to some extent, but these improvements come with increased computation and memory requirements at run-time.
We attempt to address this problem from another angle - not by changing the network structure but by altering the training method. 
We show that by temporarily pruning and then restoring a subset of the model's filters, and repeating this process cyclically, overlap in the learned features is reduced, producing improved generalization.
We show that the existing model-pruning criteria are not optimal for selecting filters to prune in this context and introduce inter-filter orthogonality as
the ranking criteria to determine under-expressive filters.
Our method is applicable both to vanilla convolutional networks and more complex modern architectures, and improves the performance across a variety of tasks, especially when applied to smaller networks.

\end{abstract}

\section{Introduction}

Convolutional Neural Networks have achieved state-of-the-art results in various computer vision tasks~\cite{He2016DeepRL, Lin2018FocalLF}. 
Much of this success is due to innovations of a novel, task-specific network architectures~\cite{He2017MaskR, Ronneberger2015UNetCN}.
Despite variation in network design, the same core optimization techniques are used across tasks.
These techniques consider each individual weight as its own entity and update them independently.
Limited progress has been made towards developing a training process specifically designed for convolutional networks, in which \textit{filters} are the fundamental unit of the network.
A filter is not a single weight parameter but a stack of spatial kernels.

Because models are typically over-parameterized, a trained convolutional network will contain redundant filters~\cite{Cogswell2015ReducingOI, Li2016PruningFF}. %
This is evident from the common practice of pruning filters~\cite{He2017ChannelPF, Anwar2017StructuredPO, Li2016PruningFF, Molchanov2016PruningCN, Liu2017LearningEC, Luo2017ThiNetAF}, rather than individual parameters~\cite{Han2015DeepCC}, to achieve model compression.
Most of these pruning methods are able to drop a significant number of filters with only a marginal loss in the performance of the model.
However, a model with fewer filters cannot be trained from scratch to achieve the performance of a large model that has been pruned to be roughly the same size~\cite{Li2016PruningFF, Luo2017ThiNetAF, Zhu2017ToPO}.
Standard training procedures tend to learn models with extraneous and prunable filters, even for architectures without any excess capacity.
This suggests that there is room for improvement in the training of Convolutional Neural Networks (ConvNets).

\begin{figure}[]
\hspace*{-2mm}
\center
    \hspace{-8.0mm}
   \includegraphics[width=0.56\linewidth]{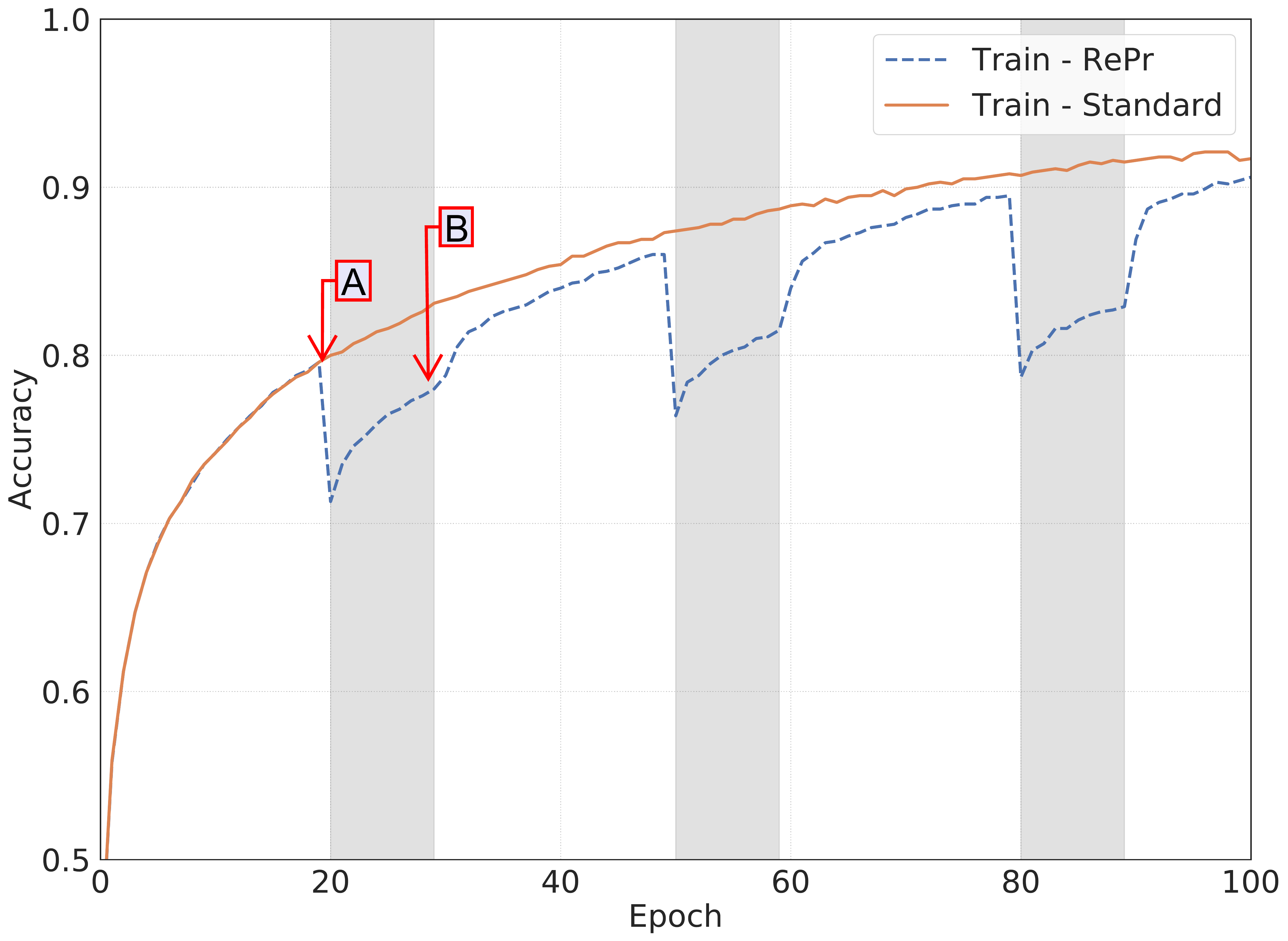}
   \hspace{-3.0mm}
   \includegraphics[width=0.535\linewidth]{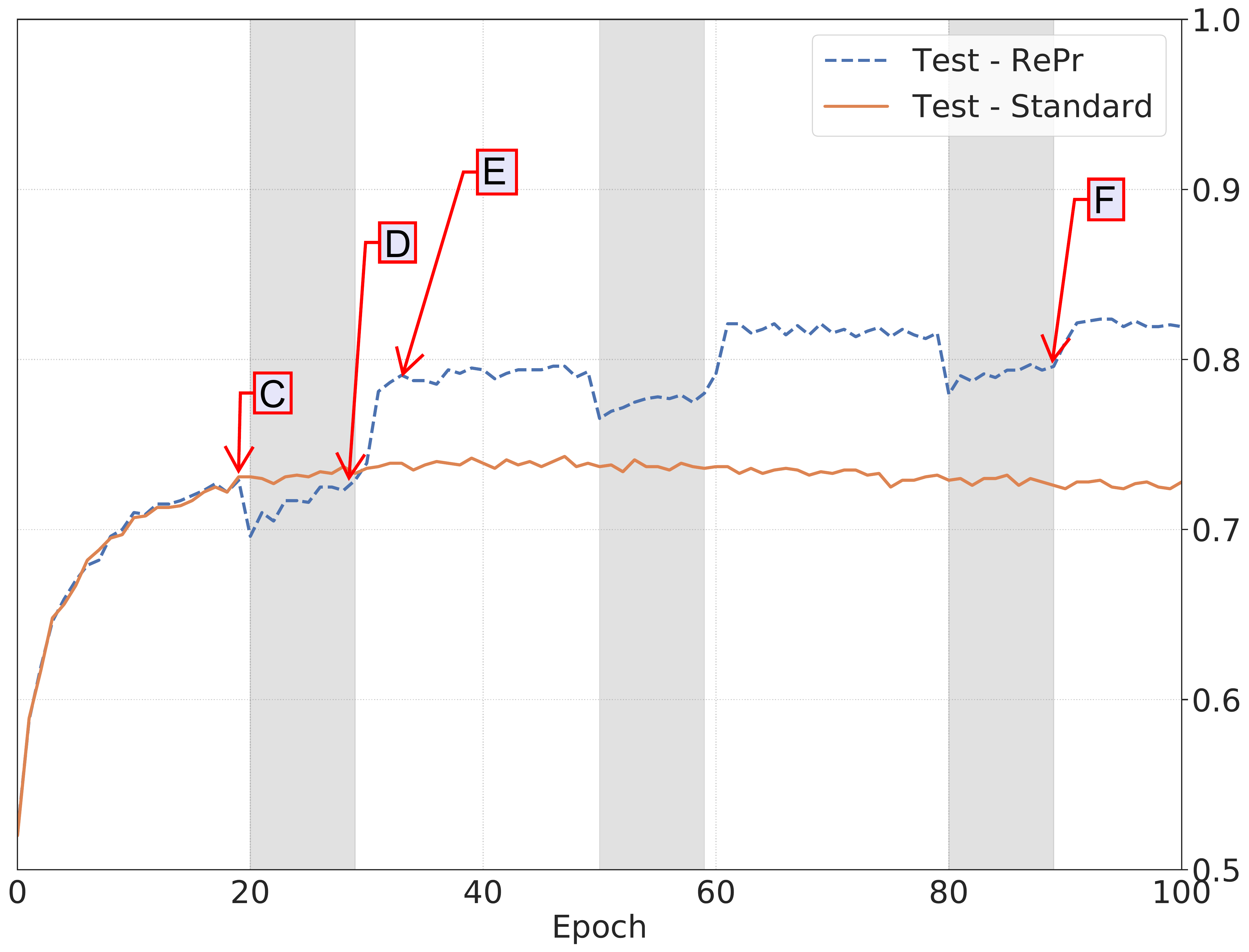}
   \hspace{-8.0mm}
   \caption{Performance of a three layer ConvNet with 32 filters each over $100$ epochs using \textcolor{orange}{standard scheme} and \textcolor{blue}{our method - RePr} on CIFAR-$10$. The shaded regions denote periods when only part of the network is trained. 
   Left: Training Accuracy, Right: Test Accuracy. Annotations [A-F] are discussed in Section~\ref{sec:training}.}
   \label{fig:REPR}
\end{figure}

To this end, we propose a training scheme in which, after some number of iterations of standard training, we select a subset of the model's filters to be temporarily dropped.
After additional training of the reduced network, we reintroduce the previously dropped filters, initialized with new weights, and continue standard training.
We observe that following the reintroduction of the dropped filters, the model is able to achieve higher performance than was obtained before the drop. 
Repeated application of this process obtains models which outperform those obtained by standard training as seen in Figure ~\ref{fig:REPR} and discussed in Section~\ref{sec:training}.
We observe this improvement across various tasks and over various types of convolutional networks.
This training procedure is able to produce improved performance across a range of possible criteria for choosing which filters to drop, and further gains can be achieved by careful selection of the ranking criterion.
According to a recent hypothesis~\cite{Frankle2018TheLT}, the relative success of over-parameterized networks may largely be due to an abundance of initial sub-networks.
Our method aims to preserve successful sub-networks while allowing the re-initialization of less useful filters.

In addition to our novel training strategy, the second major contribution of our work is an exploration of metrics to guide filter dropping.
Our experiments demonstrate that standard techniques for permanent filter pruning are suboptimal in our setting, and we present an alternative metric which can be efficiently computed, and which gives a significant improvement in performance.
We propose a metric based on the \textit{inter-}filter orthogonality within convolutional layers and show that this metric outperforms state-of-the-art filter importance ranking methods used for network pruning in the context of our training strategy.
We observe that even small, under-parameterized networks tend to learn redundant filters, which suggests that filter redundancy is not solely a result of over-parameterization, but is also due to ineffective training.
Our goal is to reduce the redundancy of the filters and increase the expressive capacity of ConvNets and we achieve this by changing the training scheme rather than the model architecture.

\section{Related Work}

\textbf{Training Scheme } Many changes to the training paradigm have been proposed to reduce over-fitting and improve generalization.
Dropout~\cite{Wu2015TowardsDT} is widely used in training deep nets. 
By stochastically dropping the neurons it prevents co-adaption of feature detectors.
A similar effect can be achieved by dropping a subset of activations ~\cite{Wan2013RegularizationON}.
Wu \etal~\cite{Wu2015TowardsDT} extend the idea of stochastic dropping to convolutional neural networks by probabilistic pooling of convolution activations.
Yet another form of stochastic training recommends randomly dropping entire layers~\cite{Huang2016DeepNW}, forcing the model to learn similar features across various layers which prevent extreme overfitting.
In contrast, our technique encourages the model to use a linear combination of features instead of duplicating the same feature.
Han \etal~\cite{Han2016DSDDT} propose Dense-Sparse-Dense (DSD), a similar training scheme, in which they apply weight regularization mid-training to encourage the development of sparse weights, and subsequently remove the regularization to restore dense weights.
While DSD works at the level of individual parameters, our method is specifically designed to apply to convolutional filters.

\textbf{Model Compression } Knowledge Distillation (KD)~\cite{Hinton2015DistillingTK} is a training scheme which uses soft logits from a larger trained model (teacher) to train a smaller model (student).
Soft logits capture hierarchical information about the object and provide a smoother loss function for optimization. This leads to easier training and better convergence for small models.
In a surprising result, Born-Again-Network~\cite{Furlanello2018BornAN} shows that if the student model is of the same capacity as the teacher it can outperform the teacher.
A few other variants of KD have been proposed~\cite{Romero2014FitNetsHF} and all of them require training several models.
Our training scheme does not depend on an external teacher and requires less training than KD. %
More importantly, when combined with KD, our method gives better performance than can be achieved by either technique independently (discussed in Section~\ref{lbl:distillation}).

\textbf{Neuron ranking } Interest in finding the least salient neurons/weights has a long history. 
LeCun~\cite{LeCun1989OptimalBD} and Hassibi\etal~\cite{Hassibi1992SecondOD} show that using the Hessian, which contains second-order derivative, identifies the weak neurons and performs better than using the magnitude of the weights.
Computing the Hessian is expensive and thus is not widely used.
Han \etal~\cite{Han2015DeepCC} show that the norm of weights is still effective ranking criteria and yields sparse models.
The sparse models do not translate to faster inference, but as a neuron ranking criterion, they are effective.
Hu \etal~\cite{Hu2016NetworkTA} explore Average Percentage of Zeros (APoZ) in the activations and use a data-driven threshold to determine the cut-off. 
Molchanov \etal~\cite{Molchanov2016PruningCN} recommend the second term from the Taylor expansion of the loss function.%
We provide detail comparison and show results on using these metrics with our training scheme in Section~\ref{sec:metric}.

\textbf{Architecture Search } %
Neural architecture search ~\cite{Liu2017ProgressiveNA, Real2018RegularizedEF, Zoph2016NeuralAS} is where the architecture is modified during training, and multiple neural network structures are explored in search of the best architecture for a given dataset. Such methods do not have any benefits if the architecture is fixed ahead of time. Our scheme improves training for a given architecture by making better use of the available parameters. This could be used in conjunction with architecture search if there is flexibility around the final architecture or used on its own when the architecture is fixed due to certified model deployment, memory requirements, or other considerations.

\textbf{Feature correlation } A well-known shortcoming of vanilla convolutional networks is their correlated feature maps~\cite{Cogswell2015ReducingOI, Glorot2010UnderstandingTD}.
Architectures like Inception-Net~\cite{Szegedy2015GoingDW} are motivated by analyzing the correlation statistics of features across layers.
They aim to reduce the correlation between the layers by using concatenated features from various sized filters, subsequent research shows otherwise~\cite{Raghu2017SVCCASV}.
More recent architectures like ResNet~\cite{He2016DeepRL} and DenseNet ~\cite{Huang2017DenselyCC} aim to implicitly reduce feature correlations by summing or concatenating activations from previous layers.
That said, these models are computationally expensive and require large memory to store previous activations.
Our aim is to induce decorrelated features without changing the architecture of the convolutional network.
This benefits all the existing implementations of ConvNet without having to change the infrastructure.
While our technique performs best with vanilla ConvNet architectures it still marginally improves the performance of modern architectures.

\section{Motivation for Orthogonal Features}
A feature for a convolutional filter is defined as the point-wise sum of the activations from individual kernels of the filter.
A feature is considered useful if it helps to improve the generalization of the model.
A model that has poor generalization usually has features that, in aggregate, capture limited directions in activation space~\cite{Morcos2017OnTI}.
On the other hand, if a model's features are orthogonal to one another, they will each capture distinct directions in activation space, leading to improved generalization.
For a trivially-sized ConvNet, we can compute the maximally expressive filters by analyzing the correlation of features across layers and clustering them into groups~\cite{Arora2014ProvableBF}.
However, this scheme is computationally impractical for the deep ConvNets used in real-world applications.
Alternatively, a computationally feasible option is the addition of a regularization term to the loss function used in standard SGD training which encourages the minimization of the covariance of the activations, but this produces only limited improvement in model performance~\cite{Rodrguez2016RegularizingCW, Cogswell2015ReducingOI}.
A similar method, in which the regularization term instead encourages the orthogonality of filter weights, has also produced marginal improvements~\cite{Brock2016NeuralPE, Poole2014AnalyzingNI, Xie2017NearOrthogonalityRI, Xie2017AllYN}.
Shang \etal~\cite{Shang2016UnderstandingAI} discovered the low-level filters are duplicated with opposite phase. 
Forcing filters to be orthogonal will minimize this duplication without changing the activation function.
In addition to improvements in performance and generalization, Saxe \etal ~\cite{Saxe2013ExactST} show that the orthogonality of weights also improves the stability of network convergence during training.
The authors of ~\cite{Xie2017AllYN, Xiao2018DynamicalIA} further demonstrate the value of orthogonal weights to the efficient training of networks.
Orthogonal initialization is common practice for Recurrent Neural Networks due to their increased sensitivity to initial conditions~\cite{Vorontsov2017OnOA}, but it has somewhat fallen out of favor for ConvNets.
These factors shape our motivation for encouraging orthogonality of features in the ConvNet and form the basis of our ranking criteria.
Because features are dependent on the input data, determining their orthogonality requires computing statistics across the entire training set, and is therefore prohibitive.  
We instead compute the orthogonality of filter weights as a surrogate.
Our experiments show that encouraging weight orthogonality through a regularization term is insufficient to promote the development of features which capture the full space of the input data manifold.
Our method of dropping overlapping filters acts as an implicit regularization and leads to the better orthogonality of filters without hampering model convergence.

We use Canonical Correlation Analysis~\cite{hotelling1936relations} (CCA) to study the overlap of features in a single layer.
CCA finds the linear combinations of random variables that show maximum correlation with each other.
It is a useful tool to determine if the learned features are overlapping in their representational capacity. 
Li \etal ~\cite{Li2015ConvergentLD} apply correlation analysis to filter activations to show that most of the well-known ConvNet architectures learn similar representations.
Raghu \etal ~\cite{Raghu2017SVCCASV} combine CCA with SVD to perform a correlation analysis of the singular values of activations from various layers.
They show that increasing the depth of a model does not always lead to a corresponding increase of the model's dimensionality, due to several layers learning representations in correlated directions.
We ask an even more elementary question - how correlated are the activations from various filters within a single layer?
In an over-parameterized network like VGG-$16$, which has several convolutional layers with $512$ filters each, it is no surprise that most of the filter activations are highly correlated.
As a result, VGG-$16$ has been shown to be easily pruned - more than $50$\% of the filters can be dropped while maintaining the performance of the full network~\cite{Molchanov2016PruningCN, Li2015ConvergentLD}.
Is this also true for significantly smaller convolutional networks, which under-fit the dataset?

\begin{figure}[]
   \includegraphics[width=0.49\linewidth]{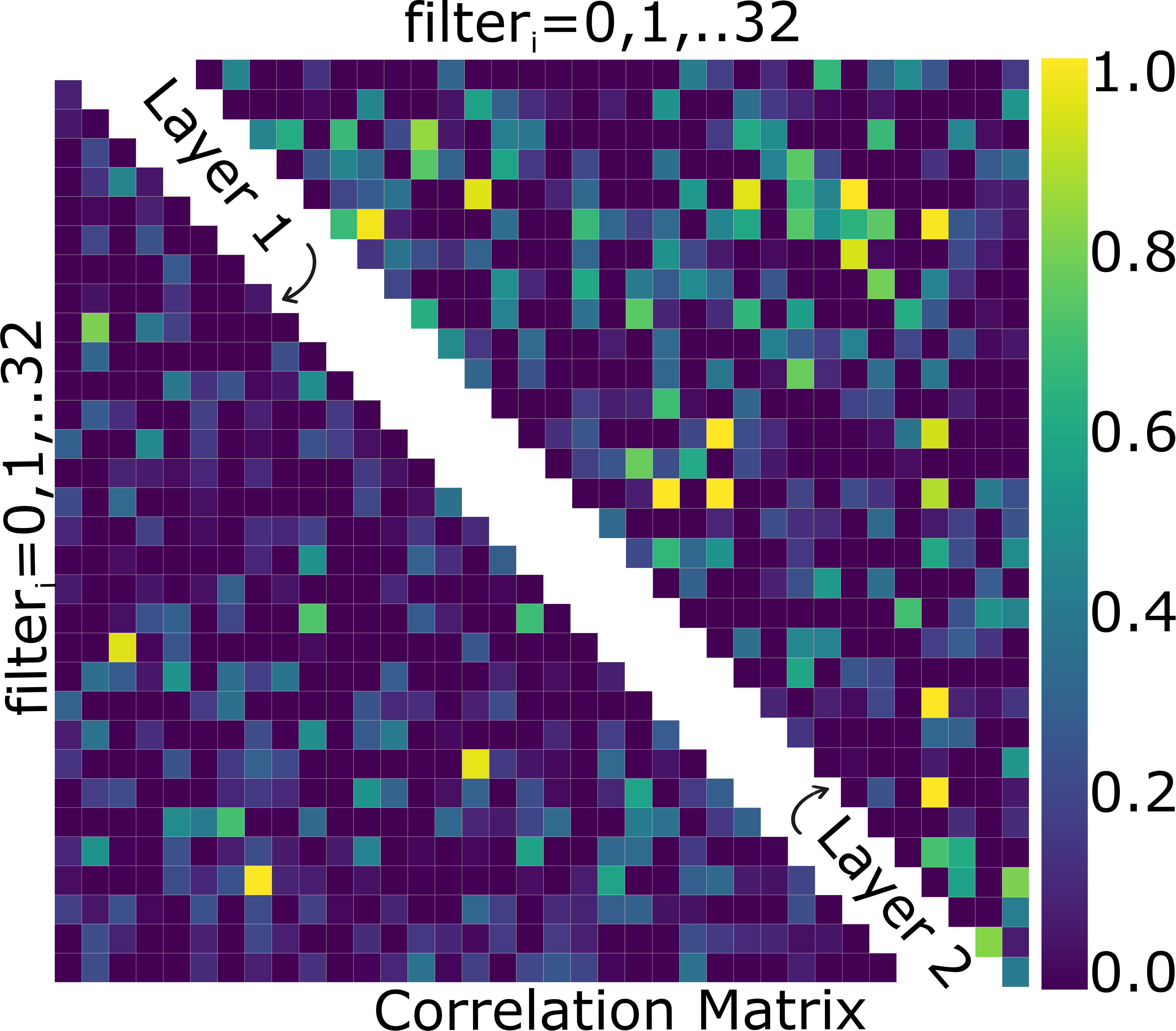}
   \includegraphics[width=0.49\linewidth]{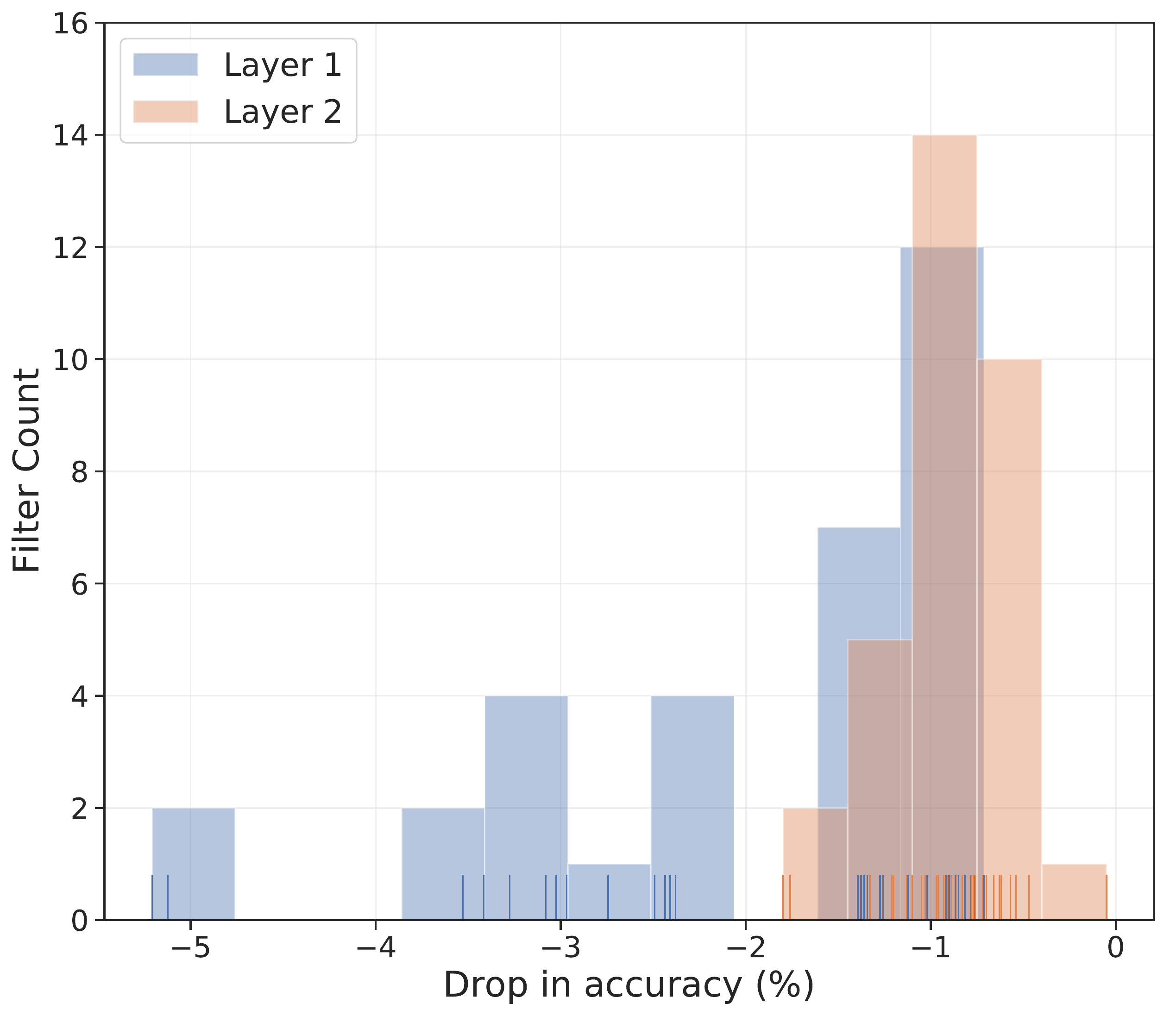}
   \caption{Left: Canonical Correlation Analysis of activations from two layers of a ConvNet trained on CIFAR-10. Right: Distribution of change in accuracy when the model is evaluated by dropping one filter at a time.}
   \label{fig:CCA}
\end{figure}

We will consider a simple network with two convolutional layers of $32$ filters each, and a softmax layer at the end.
Training this model on CIFAR-$10$ for $100$ epochs with an annealed learning rate results in test set accuracy of $58.2$\%, far below the $93.5$\% achieved by VGG-$16$.
In the case of VGG-$16$, we might expect that correlation between filters is merely an artifact of the over-parameterization of the model - the dataset simply does not have a dimensionality high enough to require every feature to be orthogonal to every other.
On the other hand, our small network has clearly failed to capture the full feature space of the training data, and thus any correlation between its filters is due to inefficiencies in training, rather than over-parameterization.

Given a trained model, we can evaluate the contribution of each filter to the model's performance by removing \textit{(zeroing out)} that filter and measuring the drop in accuracy on the test set.
We will call this metric of filter importance the \emph{"greedy Oracle"}.
We perform this evaluation independently for every filter in the model, and plot the distribution of the resulting drops in accuracy in Figure~\ref{fig:CCA} (right).
Most of the second layer filters contribute less than $1\%$ in accuracy and with first layer filters, there is a long tail. 
Some filters are important and contribute over $4\%$ of accuracy but most filters are around $1\%$. 
This implies that even a tiny and under-performing network could be filter pruned without significant performance loss.
The model has not efficiently allocated filters to capture wider representations of necessary features. 
Figure ~\ref{fig:CCA} (left) shows the correlations from linear combinations of the filter activations (CCA) at both the layers.
It is evident that in both the layers there is a significant correlation among filter activations with several of them close to a near perfect correlation of $1$ (\textit{bright yellow spots} $\color{Yellow} \blacksquare$).  The second layer (upper right diagonal) has lot more overlap of features the first layer (lower right).
For a random orthogonal matrix any value above $0.3$ (\textit{lighter than dark blue} $\color{blue} \blacksquare$) is an anomaly.
The activations are even more correlated if the linear combinations are extended to kernel functions~\cite{Hardoon2004CanonicalCA} or singular values~\cite{Raghu2017SVCCASV}.
Regardless, it suffices to say that standard training for convolutional filters does not maximize the representational potential of the network.

\section{Our Training Scheme : RePr} \label{sec:training}

We modify the training process by cyclically removing redundant filters, retraining the network, re-initializing the removed filters, and repeating. 
We consider each filter ($3$D\textit{ tensor}) as a single unit, and represent it as a long vector - ($f$).
Let $\mathbf{M}$ denote a model with $\mathcal{F}$ filters spread across $\mathbf{L}$ layers. 
Let $\mathcal{\widehat{F}}$ denote a subset of $\mathcal{F}$ filters, such that $\mathbf{M}_\mathcal{F}$ denotes a complete network whereas, $\mathbf{M}_\mathcal{F-\widehat{F}}$ denotes a sub-network without that $\mathcal{\widehat{F}}$ filters.
Our training scheme alternates between training the complete network ($\mathbf{M}_\mathcal{F}$) and the sub-network ($\mathbf{M}_\mathcal{F-\widehat{F}}$).
This introduces two hyper-parameters.
First is the number of iterations to train each of the networks before switching over; let this be $S_1$ for the full network and $S_2$ for the sub-network. 
These have to be non-trivial values so that each of the networks learns to improve upon the results of the previous network.
The second hyper-parameter is the total number of times to repeat this alternating scheme; let it be $N$. 
This value has minimal impact beyond certain range and does not require tuning.

The most important part of our algorithm is the metric used to rank the filters. 
Let $\mathcal{R}$ be the metric which associates some numeric value to a filter. 
This could be a norm of the weights or its gradients or our metric - \textit{inter-}filter orthogonality in a layer.
Here we present our algorithm agnostic to the choice of metric. Most sensible choices for filter importance results in an improvement over standard training when applied to our training scheme \textit{(see Ablation Study~\ref{sec:ablation})}.

Our training scheme operates on a macro-level and is not a weight update rule.
Thus, is not a substitute for SGD or other adaptive methods like Adam~\cite{Kingma2014AdamAM} and RmsProp~\cite{Tieleman2012}. 
Our scheme works with any of the available optimizers and shows improvement across the board. %
However, if using an optimizer that has parameters specific learning rates \textit{(like Adam)}, it is important to re-initialize the learning rates corresponding to the weights
that are part of the pruned filters ($\mathcal{\widehat{F}}$).
Corresponding Batch Normalization~\cite{Ioffe2015BatchNA} parameters ($\gamma \& \beta$) must also be re-initialized.
For this reason, comparisons of our training scheme with standard training are done with a common optimizer.

We reinitialize the filters ($\mathcal{\widehat{F}}$) to be orthogonal to its value before being dropped and the current value of non-pruned filters ($\mathcal{F-\widehat{F}}$).
We use the QR decomposition on the weights of the filters from the same layer to find the null-space and use that to find an orthogonal initialization point.

Our algorithm is training interposed with \textbf{Re}-initializing and \textbf{Pr}uning - \textbf{RePr} \textit{(pronounced: reaper)}.
We summarize our training scheme in Algorithm~\ref{alg:repr}.

\begin{algorithm}[]
    \SetKwInOut{Input}{Input}
    \SetKwInOut{Output}{Output}
    \For{$N$ iterations}{
    
        \For{$S_1$ iterations}{
            Train the full network: $\mathbf{M}_\mathcal{F}$
        }
        
        Compute the metric : $\mathcal{R}(f) \  \forall f\in \mathcal{F}$
        
        Let $\mathcal{\widehat{F}}$ be bottom $p_\%$ of $\mathcal{F}$ using $\mathcal{R}(f)$
        
        \For{$S_2$ iterations}{
            Train the sub-network : $\mathbf{M}_\mathcal{F-\widehat{F}}$
        }
        
        Reinitialize the filters ($\mathcal{\widehat{F}}$) s.t. $\mathcal{\widehat{F}} \perp \mathcal{F}$
        
        {\footnotesize \textit{(and their training specific parameters \\ from BatchNorm and Adam, if applicable)}}
        
    }
    \caption{RePr Training Scheme}
    \label{alg:repr}
\end{algorithm}

We use a shallow model to analyze the dynamics of our training scheme and its impact on the train/test accuracy. 
A shallow model will make it feasible to compute the greedy Oracle ranking for each of the filters.
This will allow us to understand the impact of training scheme alone without confounding the results due to the impact of ranking criteria.
We provide results on larger and deeper convolutional networks in Section Results~\ref{sec:results}.

Consider a $n$ layer vanilla ConvNet, without a skip or dense connections, with X filter each, as shown below:
\begin{equation*}
    \text{Img} \longmapsto \bigg[ \text{CONV}(X) \rightarrow \text{RELU}  \bigg]^n \longmapsto \text{FC} \longmapsto \text{Softmax}
\end{equation*}

We will represent this architecture as $C^n(X)$. Thus, a $C^3(32)$ has $96$ filters, and when trained with SGD with a learning rate of $0.01$, achieves test accuracy of $73\%$.
Figure~\ref{fig:REPR} shows training plots for accuracy on the training set (left) and test set (right). 
In this example, we use a RePr training scheme with $S_1=20, S_2=10, N=3, p_\%=30$ and the ranking criteria $\mathcal{R}$ as a greedy Oracle.
We exclude a separate validation set of $5$K images from the training set to compute the Oracle ranking.
In the training plot, annotation [A] shows the point at which the filters are first pruned.
Annotation [C] marks the test accuracy of the model at this point. 
The drop in test accuracy at [C] is lower than that of training accuracy at [A], which is not a surprise as most models overfit the training set.
However, the test accuracy at [D] is the same as [C] but at this point, the model only has $70\%$ of the filters. 
This is not a surprising result, as research on filter pruning shows that at lower rates of pruning most if not all of the performance can be recovered~\cite{Molchanov2016PruningCN}.

What is surprising is that test accuracy at [E], which is only a couple of epochs after re-introducing the pruned filters, is significantly higher than point [C].
Both point [C] and point [E] are same capacity networks, and higher accuracy at [E] is not due to the model convergence. 
In the standard training (\textcolor{orange}{orange line}) the test accuracy does not change during this period.
Models that first grow the network and then prune~\cite{Dai2017NeSTAN, Han2015LearningBW}, unfortunately, stopped shy of another phase of growth, which yields improved performance. 
In their defense, this technique defeats the purpose of obtaining a smaller network by pruning. 
However, if we continue RePr training for another two iterations, we see that the point [F], which is still at $70\%$ of the original filters yields accuracy which is comparable
to the point [E] ($100\%$ of the model size.

Another observation we can make from the plots is that training accuracy of RePr model is lower, which signifies some form of regularization on the model. 
This is evident in the Figure~\ref{fig:ablation_percent} (Right), which shows RePr with a large number of iterations ($N=28$). 
While the marginal benefit of higher test accuracy diminishes quickly, the generalization gap between train and test accuracy is reduced significantly.

\section{Our Metric : \textit{inter}-filter orthogonality} \label{sec:metric}

The goals of searching for a metric to rank least important filters are twofold - (1) computing the greedy Oracle is not computationally feasible for large networks, and (2) the greedy Oracle may not be the best criteria. If a filter which captures a unique direction, thus not replaceable by a linear combination of other filters, has a lower contribution to accuracy, the Oracle will drop that filter. On a subsequent re-initialization and training, we may not get back the same set of directions.

The directions captured by the activation pattern expresses the capacity of a deep network~\cite{Raghu2017OnTE}. 
Making orthogonal features will maximize the directions captured and thus expressiveness of the network.
In a densely connected layer, orthogonal weights lead to orthogonal features, even in the presence of ReLU~\cite{Vorontsov2017OnOA}.
However, it is not clear how to compute the orthogonality of a convolutional layer. 

A convolutional layer is composed of parameters grouped into spatial kernels and sparsely share the incoming activations.
Should all the parameters in a single convolutional layer be considered while accounting for orthogonality?
The theory that promotes initializing weights to be orthogonal is based on densely connected layers (FC-layers) and popular deep learning libraries follow this guide\footnote{tensorflow:ops/init\_ops.py\#L543 \& pytorch:nn/init.py\#L350} by considering convolutional layer as one giant vector disregarding the sparse connectivity.
A recent attempt to study orthogonality of convolutional filters is described in ~\cite{Xiao2018DynamicalIA} but their motivation is the convergence of very deep networks (10K layers) and not orthogonality of the features.
Our empirical study suggests a strong preference for requiring orthogonality of individual filters in a layer (inter-filter \& intra-layer) rather than individual kernels.

A filter of kernel size $k\times k$ is commonly a $3$D tensor of shape $k \times k \times c$, where $c$ is the number of channels in the incoming activations.
Flatten this tensor to a $1$D vector of size $k*k*c$, and denote it by $f$.
Let $J_\ell$ denote the number of filters in the layer $\ell$, where $\ell \in \mathbf{L}$, and $\mathbf{L}$ is the number of layers in the ConvNet.
Let $\boldsymbol{W}_\ell$ be a matrix, such that the individual rows are the flattened filters ($f$) of the layer $\ell$.

Let $\boldsymbol{\hat{W}_\ell} = \boldsymbol{W_\ell}/||\boldsymbol{W_\ell}||$ denote the normalized weights.
Then, the measure of Orthogonality for filter $f$ in a layer $\ell$ (denoted by $O_\ell^f$) is computed as shown in the equations below.

\begin{equation}
\boldsymbol{P}_\ell = |\boldsymbol{\hat{W}_\ell} \times \boldsymbol{\hat{W}_\ell}^T - I |
\end{equation}

\begin{equation}
\label{eqn:ortho}
O^f_\ell = \frac{\sum \boldsymbol{P_\ell}[f]}{J_\ell}
\end{equation}

$\boldsymbol{P}_\ell$ is a matrix of size $J_\ell \times J_\ell$ and $\boldsymbol{P}[i]$ denotes $i^{\textit{th}}$ row of $\boldsymbol{P}$.
Off-diagonal elements of a row of $\boldsymbol{P}$ for a filter $f$ denote projection of all the other filters in the same layer with $f$.
The sum of a row is minimum when other filters are orthogonal to this given filter.
We rank the filters least important (thus subject to pruning) if this value is largest among all the filters in the network.
While we compute the metric for a filter over a single layer, the ranking is computed over all the filters in the network.
We do not enforce per layer rank because that would require learning a hyper-parameter $p_\%$ for every layer and some layers are more sensitive than others.
Our method prunes more filters from deeper layers compared to the earlier layers.
This is in accordance with the distribution of contribution of each filter in a given network (Figure~\ref{fig:CCA} right).

Computation of our metric does not require expensive calculations of the inverse of Hessian~\cite{LeCun1989OptimalBD} or the second order derivatives~\cite{Hassibi1992SecondOD} and is feasible for any sized networks. 
The most expensive calculations are $L$ matrix products of size $J_\ell \times J_\ell$, but GPUs are designed for fast matrix-multiplications. 
Still, our method is more expensive than computing norm of the weights or the activations or the Average Percentage of Zeros (APoZ).

Given the choice of Orthogonality of filters, an obvious question would be to ask if adding a soft penalty to the loss function improve this training?
A few researchers  ~\cite{Brock2016NeuralPE, Poole2014AnalyzingNI, Xie2017NearOrthogonalityRI} have reported marginal improvements due to added regularization in the ConvNets used for task-specific models.
We experimented by adding $\lambda * \sum_\ell \boldsymbol{P}_\ell$ to the loss function, but we did not see any improvement. 
Soft regularization penalizes all the filters and changes the loss surface to encourage random orthogonality in the weights without improving expressiveness.

\begin{figure}
\begin{minipage}{0.7\linewidth}
  \begin{minipage}[c]{0.25\linewidth}
    \centering
    \includegraphics[width=4.0\linewidth]{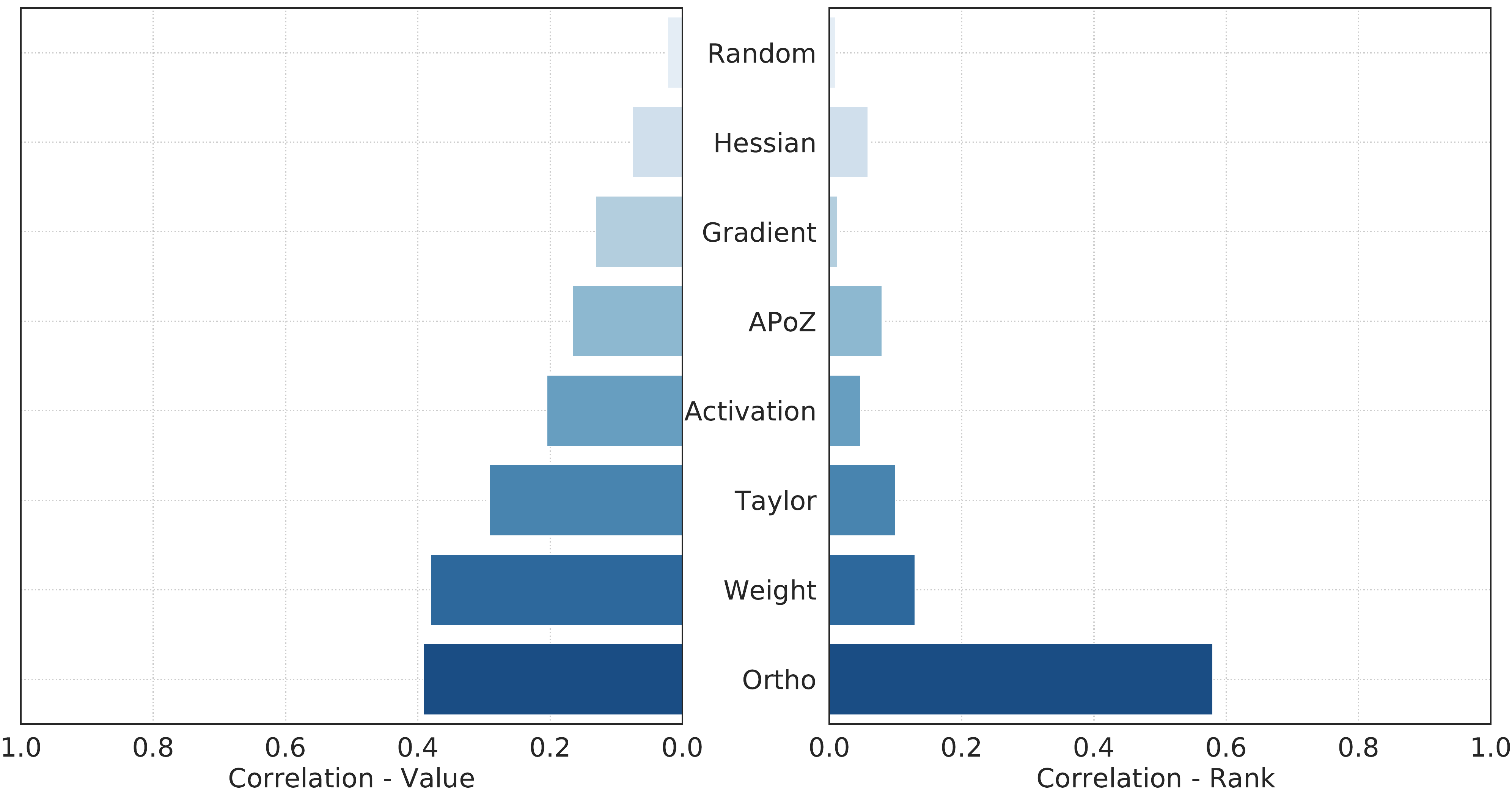}
  \end{minipage}
  \hfill
  \begin{minipage}[c]{0.01\linewidth}
    \scriptsize
\begin{tabular}{rl}
                   \multicolumn{2}{c}{\textbf{\shortstack{CIFAR-10}} - $C^3(32)$} \\ \hline
                  \toprule
                  Standard             & 72.1                                  \\ \hdashline
                  \rowcolor[HTML]{EFEFEF} 
                  Random          & 73.4                                  \\
                  Activations          & 74.1                                  \\
                  \rowcolor[HTML]{EFEFEF} 
                  APoZ~\cite{Hu2016NetworkTA}                 & 74.3                                \\
                  Gradients~\cite{LeCun1989OptimalBD}            & 74.3                                  \\
                  \rowcolor[HTML]{EFEFEF} 
                  Taylor~\cite{Molchanov2016PruningCN}               & 74.3                                  \\
                  Hessian~\cite{LeCun1989OptimalBD}              & 74.4                                  \\
                  \rowcolor[HTML]{EFEFEF} 
                  Weights~\cite{Han2015DeepCC}              & 74.6                                  \\
                  Oracle               & 76.0                                  \\
                  \rowcolor[HTML]{EFEFEF} 
                  \textcolor{green}{Ortho}                & \textbf{76.4}                                  \\ \hline
    \end{tabular}
    \end{minipage}
  \end{minipage}
    \caption{Left: Pearson correlation coefficient of various metric values with the accuracy values from greedy Oracle.
   Center: Pearson correlation coefficient of filter ranks using various metric with rank from greedy Oracle 
  Right: Test accuracy on CIFAR-10 using standard training and RePr training with various metrics}
  \label{fig:corr}
\end{figure}

\section{Ablation study} \label{sec:ablation}

\begin{figure}[]
\center
    \includegraphics[width=0.5\linewidth]{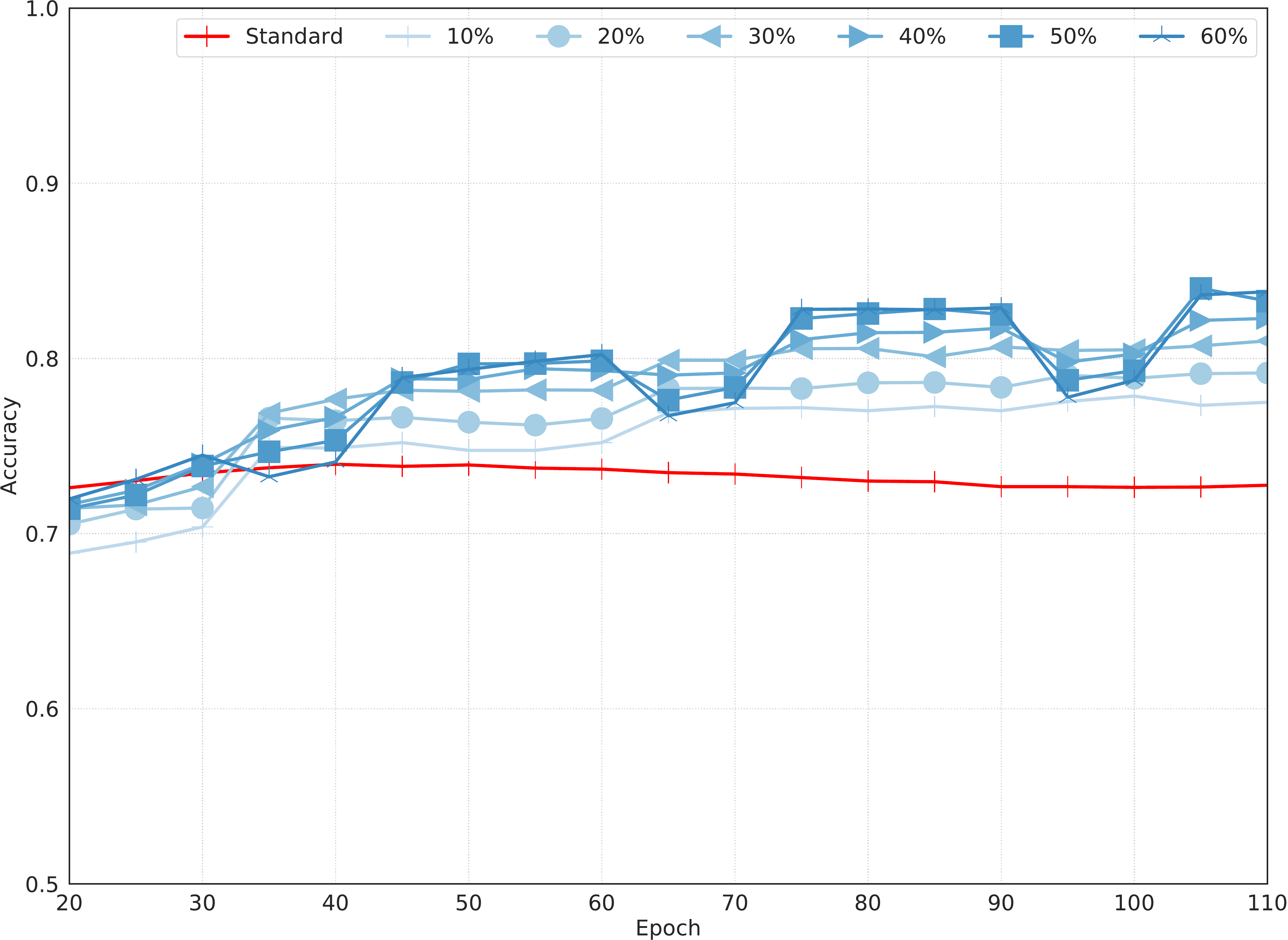}
    \hspace{-0.1cm}\includegraphics[width=0.48\linewidth]{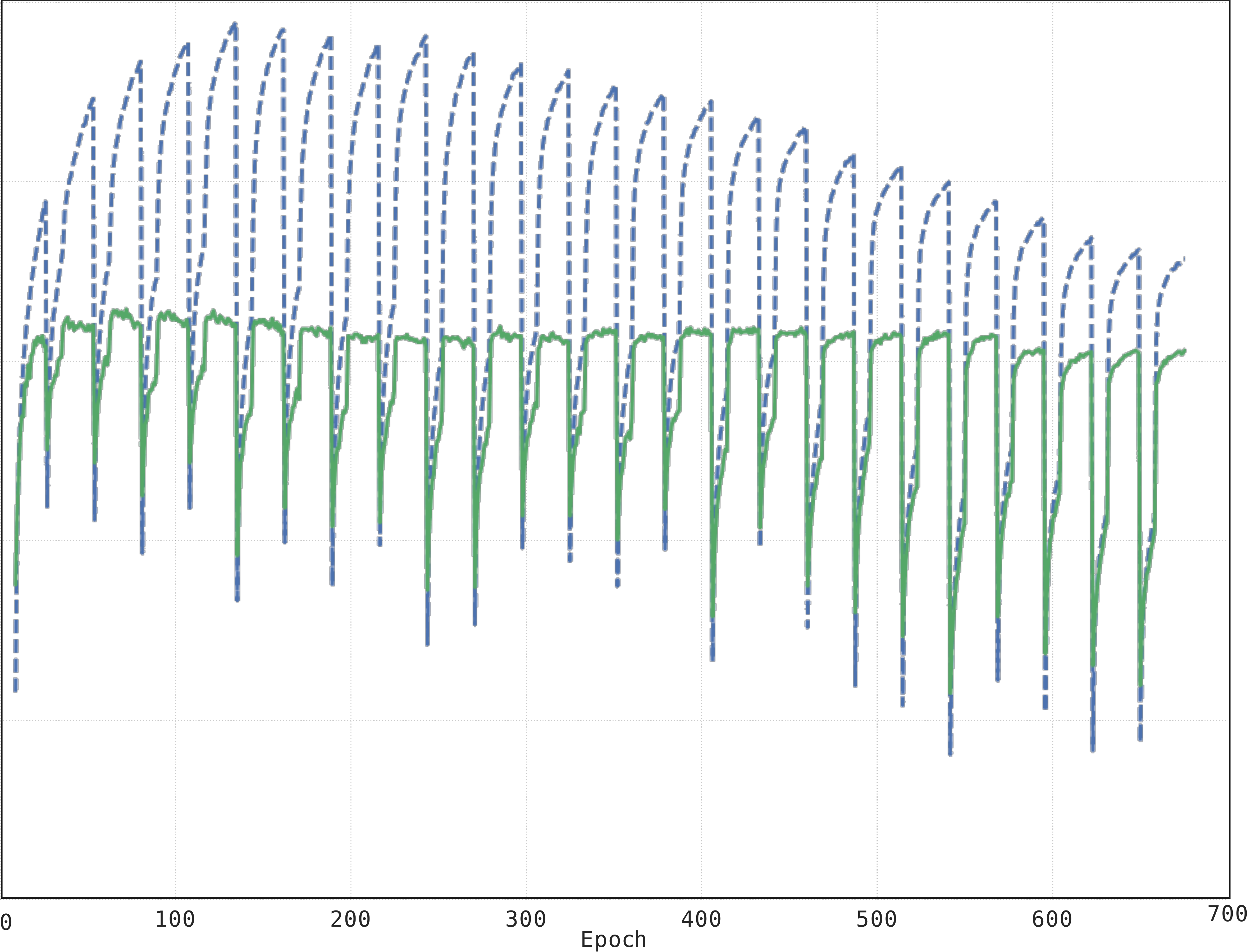}
   \caption{Left: RePr training with various percentage of filters pruned. Shows average test accuracy over 5 epochs starting from epoch 20 for better visibility.
   Right:Marginal returns of multiple iterations of RePR - \textcolor{blue}{Training} and \textcolor{green}{Test} accuracy on CIFAR-10}
   \label{fig:ablation_percent}
\end{figure}

\textbf{Comparison of pruning criteria }
We measure the correlation of our metric with the Oracle to answer the question - how good a substitute is our metric for the filter importance ranking. 
Pearson correlation of our metric, henceforth referred to as Ortho, with the Oracle is $0.38$. 
This is not a strong correlation, however, when we compare this with other known metrics, it is the closest.
Molchanov \etal ~\cite{Molchanov2016PruningCN} report Spearman correlation of their criteria (Taylor) with greedy Oracle at $0.73$. 
We observed similar numbers for Taylor ranking during the early epochs but the correlation diminished significantly as the models converged.
This is due to low gradient value from filters that have converged.
The Taylor metric is a product of the activation and the gradient. 
High gradients correlate with important filters during early phases of learning but when models converge low gradient do not necessarily mean less salient weights.
It could be that the filter has already converged to a useful feature that is not contributing to the overall error of the model or is stuck at a saddle point.
With the norm of activations, the relationship is reversed. 
Thus by multiplying the terms together hope is to achieve a balance. 
But our experiments show that in a fully converged model, low gradients dominate high activations.
Therefore, the Taylor term will have lower values as the models converge and will no longer be correlated with the inefficient filters.
While the correlation of the values denotes how well the metric is the substitute for predicting the accuracy, it is more important to measure the correlation of the rank of the filters.
Correlation of the values and the rank may not be the same, and the correlation with the rank is the more meaningful measurement to determine the weaker filters.
Ortho has a correlation of $0.58$ against the Oracle when measured over the rank of the filters. 
Other metrics show very poor correlation using the rank. Figure~\ref{fig:corr} (Left and Center) shows the correlation plot for various metrics with the Oracle.
The table on the right of Figure~\ref{fig:corr} presents the test accuracy on CIFAR-10 of various ranking metrics.
From the table, it is evident that Orthogonality ranking leads to a significant boost of accuracy compared to standard training and other ranking criteria.

\textbf{Percentage of filters pruned }
One of the key factors in our training scheme is the percentage of the filters to prune at each pruning phase ($p_\%$). It behaves like the Dropout parameter, and impacts the training time and generalization ability of the model \textit{(see Figure:~\ref{fig:ablation_percent})}. 
In general the higher the pruned percentage, the better the performance. However, beyond $30\%$, the performances are not significant.
Up to $50\%$, the model seems to recover from the dropping of filters. 
Beyond that, the training is not stable, and sometimes the model fails to converge.

\textbf{Number of RePr iterations }
Our experiments suggest that each repeat of the RePr process has diminishing returns, and therefore should be limited to a single-digit number (see Figure~\ref{fig:ablation_percent} (Right)).
Similar to Dense-Sparse-Dense~\cite{Han2016DSDDT} and Born-Again-Networks~\cite{Furlanello2018BornAN}, we observe that for most networks, two to three iterations is sufficient to achieve the maximum benefit.

\begin{figure}[]
   \includegraphics[width=1.0\linewidth]{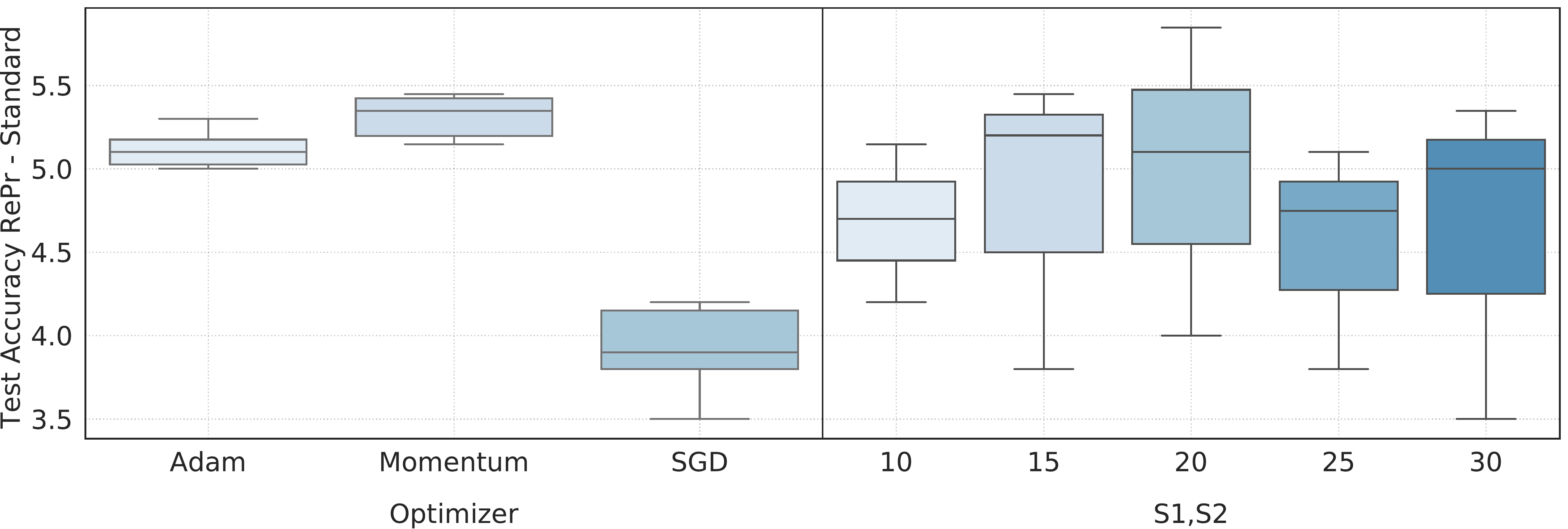}
   \caption{Left: Impact of using various optimizers on RePr training scheme. Right: Results from using different S1/S2 values. For clarity, these experiments only shows results with $S1=S2$}
   \label{fig:ablation_optimizer_s1s2}
\end{figure}

\textbf{Optimizer and S1/S2 }
Figure~\ref{fig:ablation_optimizer_s1s2} (left) shows variance in improvement when using different optimizers. 
Our model works well with most well-known optimizers. Adam and Momentum perform better than SGD due to their added stability in training. 
We experimented with various values of $S1$ and $S2$, and there is not much difference if either of them is large enough for the model to converge temporarily. 

\begin{figure}[]
\hspace*{-4.0mm}
\center
   \includegraphics[width=0.50\linewidth]{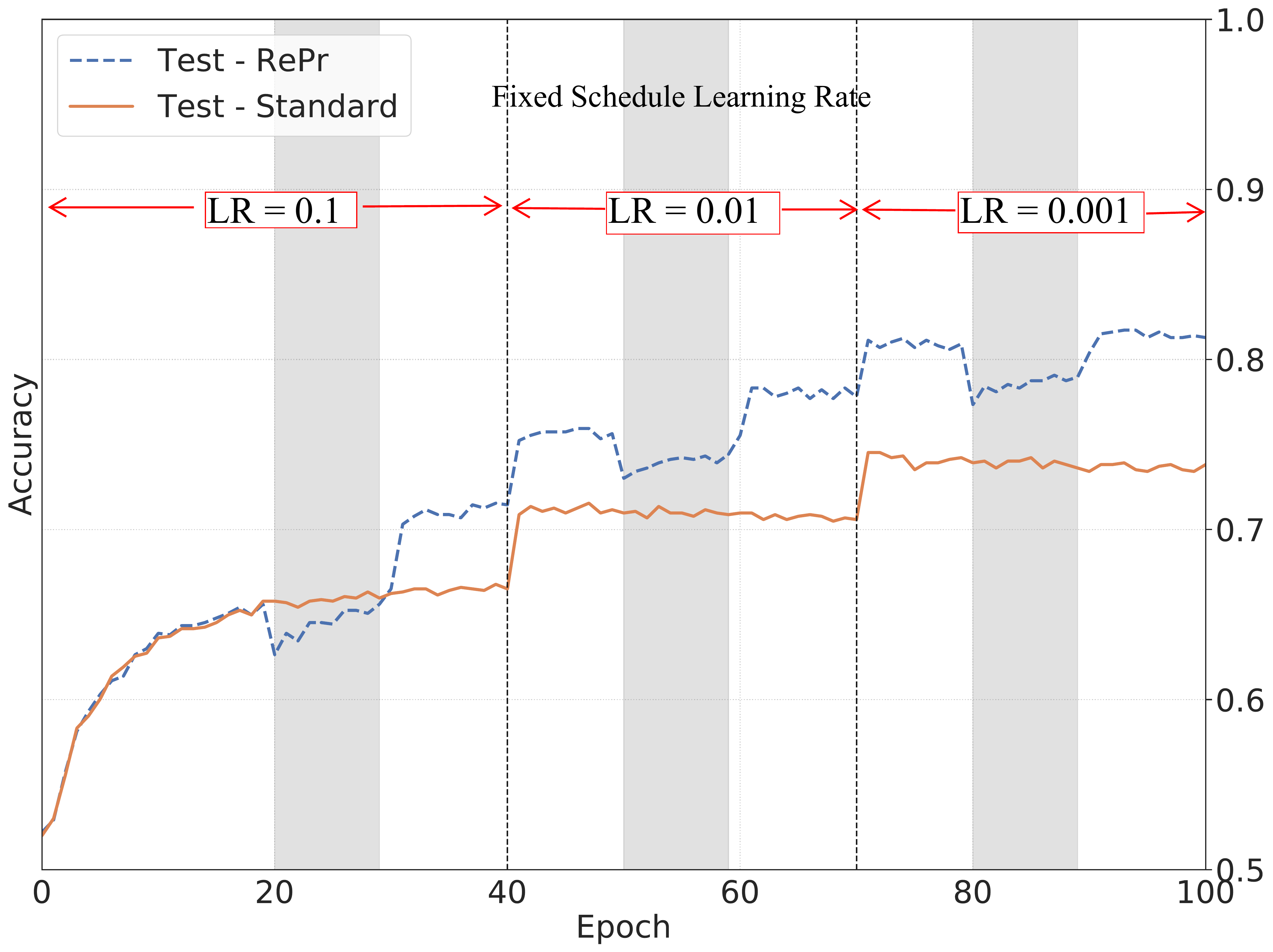}
   \hspace{-2.0mm}
   \includegraphics[width=0.50\linewidth]{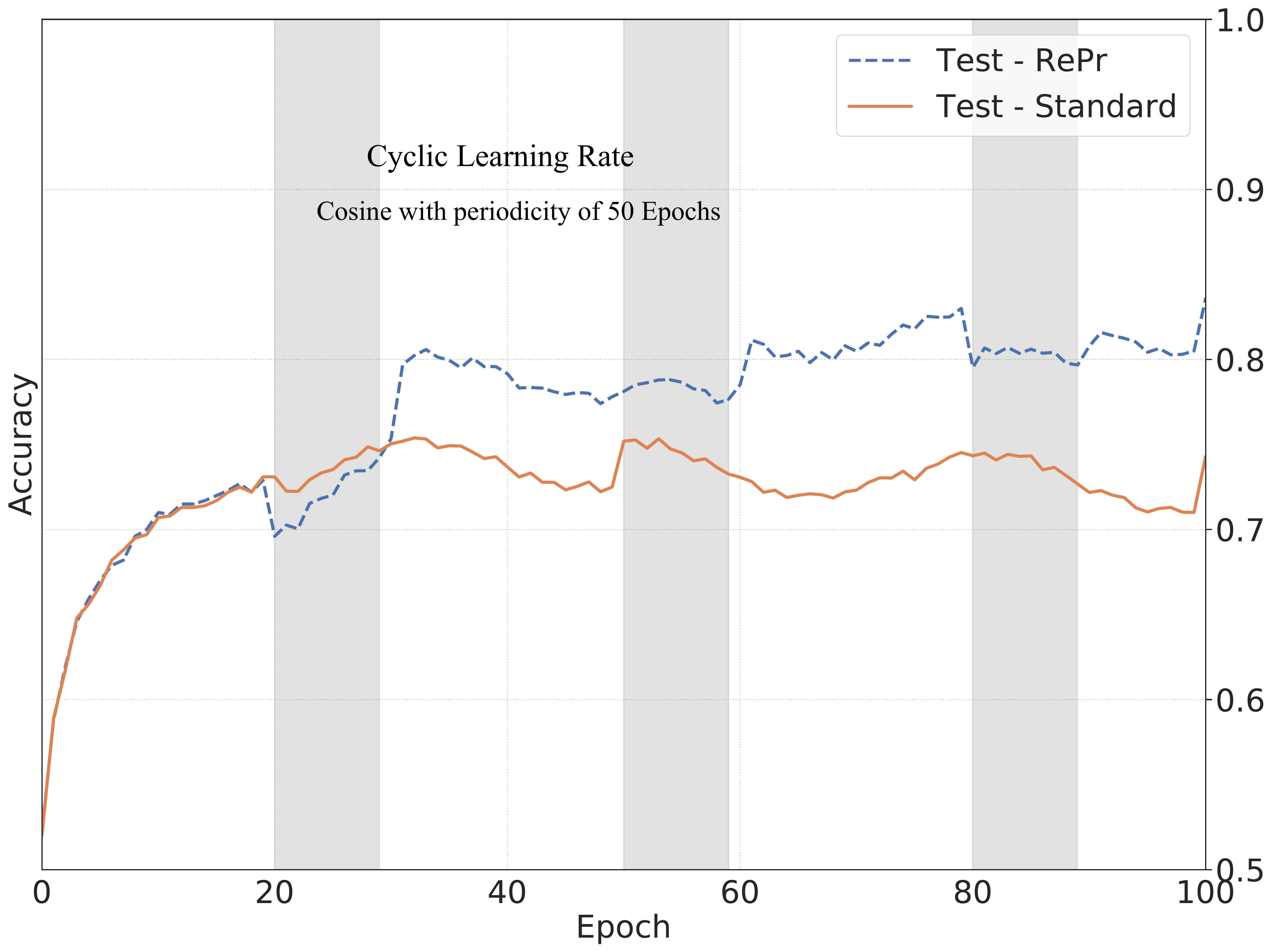}
   \caption{Test accuracy of a three layer ConvNet with 32 filters each over $100$ epochs using \textcolor{orange}{standard scheme} and \textcolor{blue}{our method - RePr} on CIFAR-$10$. The shaded regions denote periods when only part of the network is trained for RePr.
   Left: Fixed Learning Rate schedule of $0.1$, $0.01$ and $0.001$.Right: Cyclic Learning Rate with periodicity of $50$ Epochs, and amplitude of 0.005 and starting LR of 0.001.}
   \label{fig:LR}
\end{figure}

\textbf{Learning Rate Schedules}
SGD with a fixed learning rate does not typically produce optimal model performance.
Instead, gradually annealing the learning rate over the course of training is known to produce models with higher test accuracy.
State-of-the-art results on ResNet, DenseNet, Inception were all reported with a predetermined learning rate schedule.
However, the selection of the exact learning rate schedule is itself a hyperparameter, one which needs to be specifically tuned for each model.
Cyclical learning rates~\cite{Smith2017CyclicalLR} can provide stronger performance without exhaustive tuning of a precise learning rate schedule.
Figure~\ref{fig:LR} shows the comparison of our training technique when applied in conjunction with fixed schedule learning rate scheme and cyclical learning rate. 
Our training scheme is not impacted by using these schemes, and improvements over standard training is still apparent.

\textbf{Impact of Dropout}
Dropout, while commonly applied in Multilayer Perceptrons, is typically not used for ConvNets.
Our technique can be viewed as a type of non-random Dropout, specifically applicable to ConvNets.
Unlike standard Dropout, out method acts on entire filters, rather than individual weights, and is applied only during select stages of training, rather than in every training step.
Dropout prevents overfitting by encouraging co-adaptation of weights.
This is effective in the case of over-parameterized models, but in compact or shallow models, Dropout may needlessly reduce already limited model capacity.

\begin{figure}[]
\center
   \includegraphics[width=0.48\linewidth]{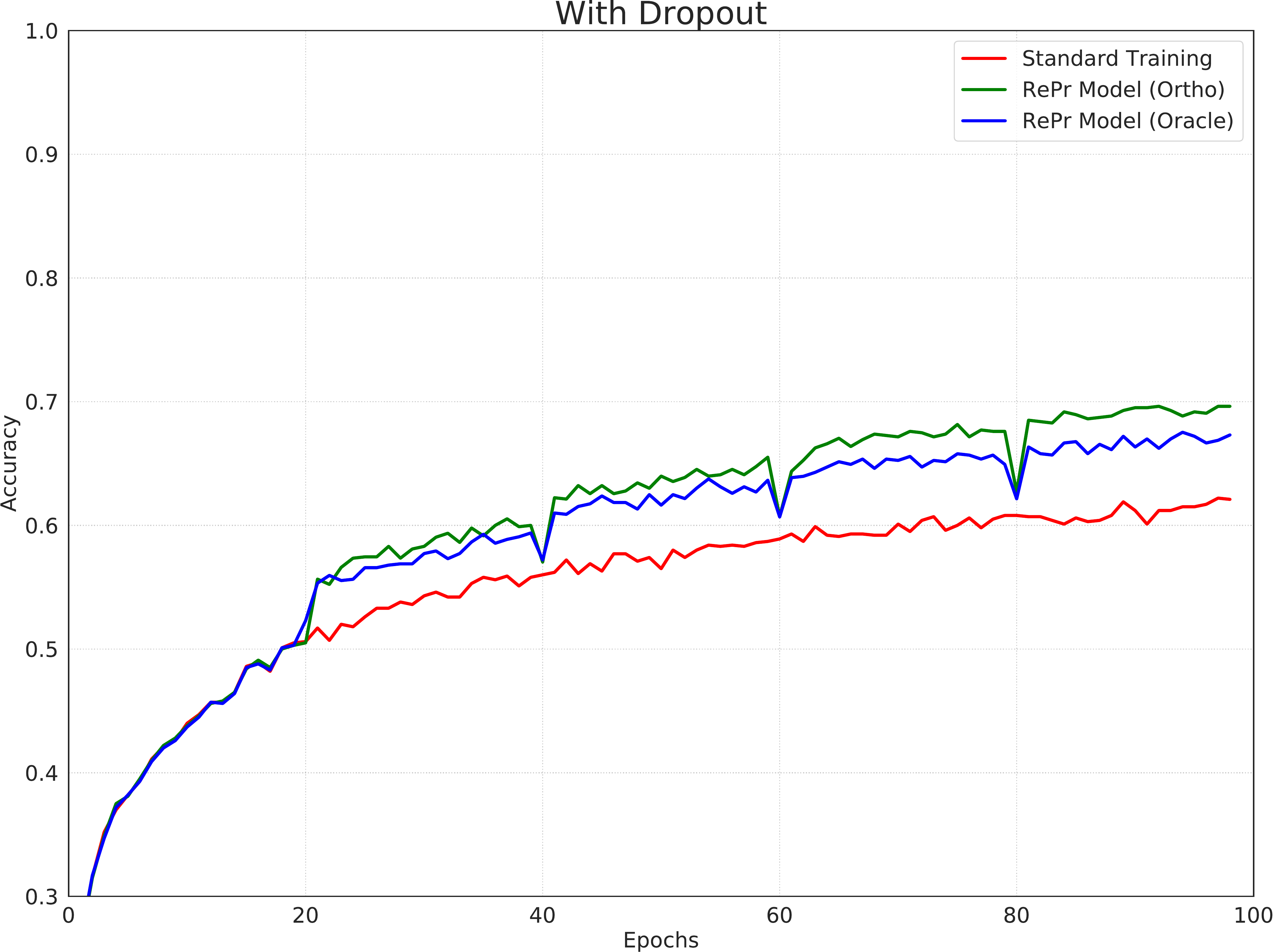}
   \includegraphics[width=0.48\linewidth]{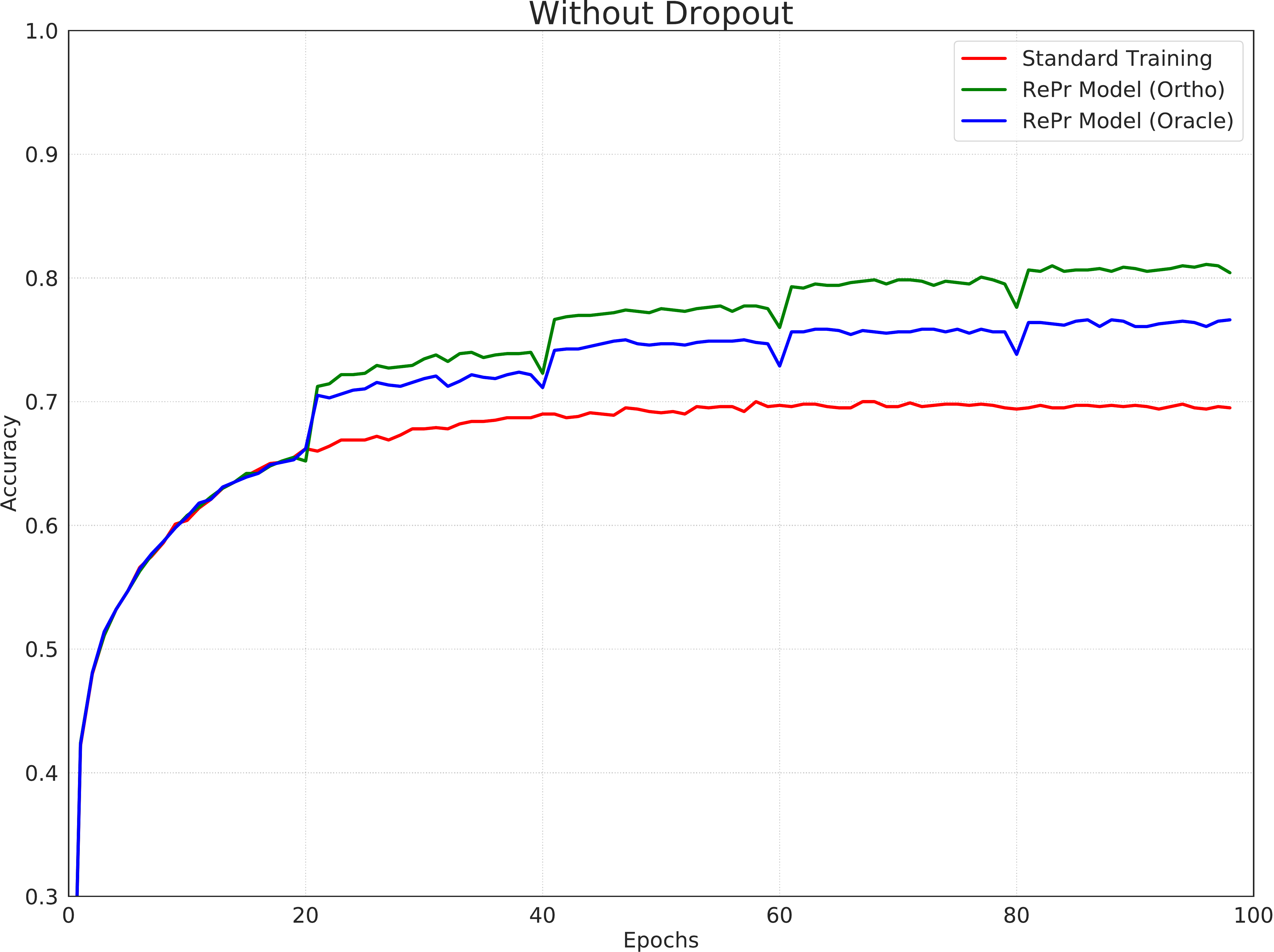}
   \caption{Test accuracy of a three layer ConvNet with 32 filters each over $100$ epochs using \textcolor{red}{standard scheme}, \textcolor{blue}{RePr with Oracle} and \textcolor{green}{RePr with Ortho} on CIFAR-$10$. Left: With Dropout of $0.5$. Right: No Dropout}
   \label{fig:dropout}
\end{figure}

Figure~\ref{fig:dropout} shows the performance of Standard Training and our proposed method (RePr) with and without Dropout on a three-layer convolutional neural network with $32$ filters each. 
Dropout was applied with a probability of $0.5$.
We observe that the inclusion of Dropout lowers the final test accuracy, due to the effective reduction of the model's capacity by half.
Our method produces improved performance with or without the addition of standard Dropout, demonstrating that its effects are distinct from the benefits of Dropout. 

\textbf{Orthogonal Loss - OL}
Adding Orthogonality of filters (equation 1) as a regularization term as a part of the optimization loss does not significantly impact the performance of the model.
Thus, the loss function will be -
\begin{equation*}
    \mathcal{L} = \text{Cross entropy} + \lambda * |\boldsymbol{\hat{W}_\ell} \times \boldsymbol{\hat{W}_\ell}^T - I |
\end{equation*}
where, $\lambda$ is a hyper-parameter which balances both the cost terms. We experimented with various values of $\lambda$. Table~\ref{tbl:ortho_loss} report the results with this loss term for the $\lambda = 0.01$, for which the validation accuracy was the highest. OL refers to addition of this loss term.

\begin{table}[H]
\center
\begin{tabular}{ccccc}
\toprule
          $C^3(32)$ & \multicolumn{1}{c}{\textbf{Std}} & \multicolumn{1}{c}{\textbf{Std+OL}} & \multicolumn{1}{c}{\textbf{RePr}} & \multicolumn{1}{c}{\textbf{RePr+OL}} \\ \hline
CIFAR-10  & 72.1                                                     & 72.8                                                    & 76.4                                                      & {\color[HTML]{3166FF} 76.7}                                                         \\
CIFAR-100 & 47.2                                                     & 48.3                                                    & 58.2                                                      & {\color[HTML]{3166FF} 58.6}                                                        \\ \bottomrule
\end{tabular}
\caption{Comparison of addition of Orthogonality loss to Standard Training and RePr}
\label{tbl:ortho_loss}
\end{table}

\begin{figure}[]
   \includegraphics[width=1.0\linewidth]{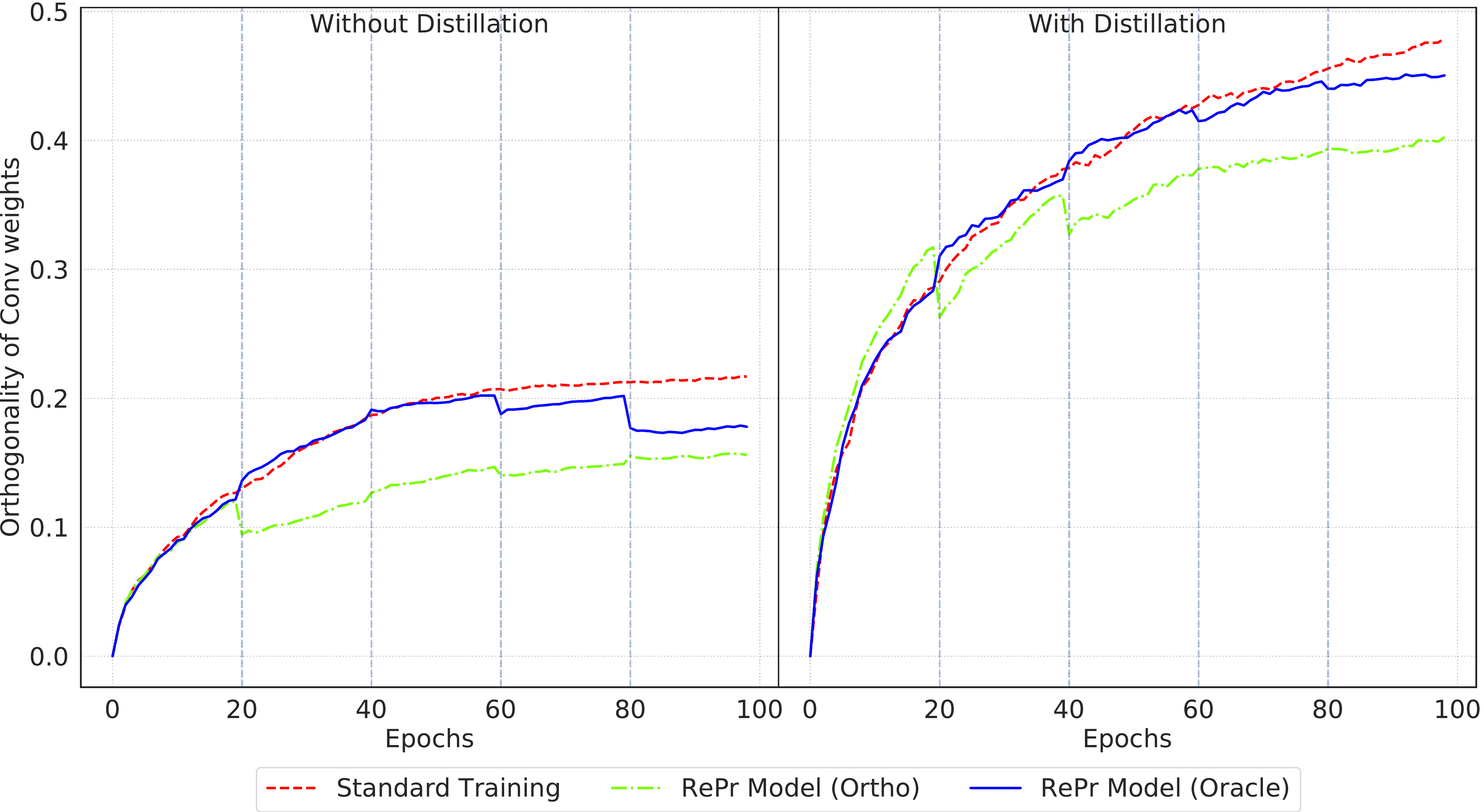}
   \caption{Comparison of orthogonality of filters (Ortho-sum - eq 2) in standard training and RePr training with and without Knowledge Distillation. \textbf{Lower value} signifies less overlapping filters.
   Dashed vertical lines denotes filter dropping.}
   \label{fig:orthodistill}
\end{figure}

\begin{table}[]
\center
\begin{tabular}{
>{\columncolor[HTML]{FFFFFF}}c 
>{\columncolor[HTML]{FFFFFF}}c 
>{\columncolor[HTML]{FFFFFF}}c
>{\columncolor[HTML]{FFFFFF}}c 
>{\columncolor[HTML]{FFFFFF}}c }
\toprule
          $C^3(32)$ & \multicolumn{1}{c}{\cellcolor[HTML]{FFFFFF}\textbf{Std}} & \multicolumn{1}{c}{\cellcolor[HTML]{FFFFFF}\textbf{KD}} & \multicolumn{1}{c}{\cellcolor[HTML]{FFFFFF}\textbf{RePr}} & \multicolumn{1}{c}{\cellcolor[HTML]{FFFFFF}\textbf{KD+RePr}} \\ \hline
CIFAR-10  & 72.1                                                     & 74.8                                                    & 76.4                                                      & \textbf{83.1}                                                         \\
CIFAR-100 & 47.2                                                     & 56.5                                                    & 58.2                                                      & \textbf{64.1}                                                        \\ \bottomrule
\end{tabular}
\caption{Comparison of Knowledge Distillation with RePr.}
\label{tbl:distillation}
\end{table}

\section{Orthogonality and Distillation} \label{lbl:distillation}
Our method, RePr and Knowledge Distillation (KD) are both techniques to improve performance of compact models.
RePr reduces the overlap of filter representations and KD distills the information from a larger network. 
We present a brief comparison of the techniques and show that they can be combined to achieve even better performance.

RePr repetitively drops the filters with most overlap in the directions of the weights using the \textit{inter}-filter orthogonality, as shown in the equation~\ref{eqn:ortho}. 
Therefore, we expect this value to gradually reduce over time during training.
Figure~\ref{fig:orthodistill} (left) shows the sum of this value over the entire network with three training schemes.
We show RePr with two different filter ranking criteria - \textcolor{green}{Ortho} and \textcolor{blue}{Oracle}. 
It is not surprising that RePr training scheme with Ortho ranking has lowest Ortho sum but it is surprising that RePr training with Oracle ranking also reduces the filter overlap, compared to the standard training.
Once the model starts to converge, the least important filters based on Oracle ranking are the ones with the most overlap.
And dropping these filters leads to better test accuracy \textit{(table on the right of  Figure~\ref{fig:corr}).}
Does this improvement come from the same source as the that due to Knowledge Distillation?
Knowledge Distillation (KD) is a well-proven methodology to train compact models. 
Using soft logits from the teacher and the ground truth signal the model converges to better optima compared to standard training. 
If we apply KD to the same three experiments (see Figure~\ref{fig:orthodistill}, right), we see that all the models have significantly larger Ortho sum. Even the RePr (Ortho) model struggles to lower the sum as the model is strongly guided to converge to a specific solution.
This suggests that this improvement due to KD is not due to reducing filter overlap. Therefore, a model which uses both the techniques should benefit by even better generalization.
Indeed, that is the case as the combined model has significantly better performance than either of the individual models, as shown in  Table~\ref{tbl:distillation}. 

\section{Results} \label{sec:results}
We present the performance of our training scheme, RePr, with our ranking criteria, \textit{inter}-filter orthogonality, Ortho, on different ConvNets~\cite{Simonyan2014VeryDC, He2016DeepRL, Szegedy2015GoingDW, Szegedy2016RethinkingTI, Huang2017DenselyCC}.
For all the results provided RePr parameters are: $S_1=20$, $S_2=10$, $p_\%=30$, and with three iterations, $N=3$.

\begin{figure}[]
\center
   \includegraphics[width=1.0\linewidth]{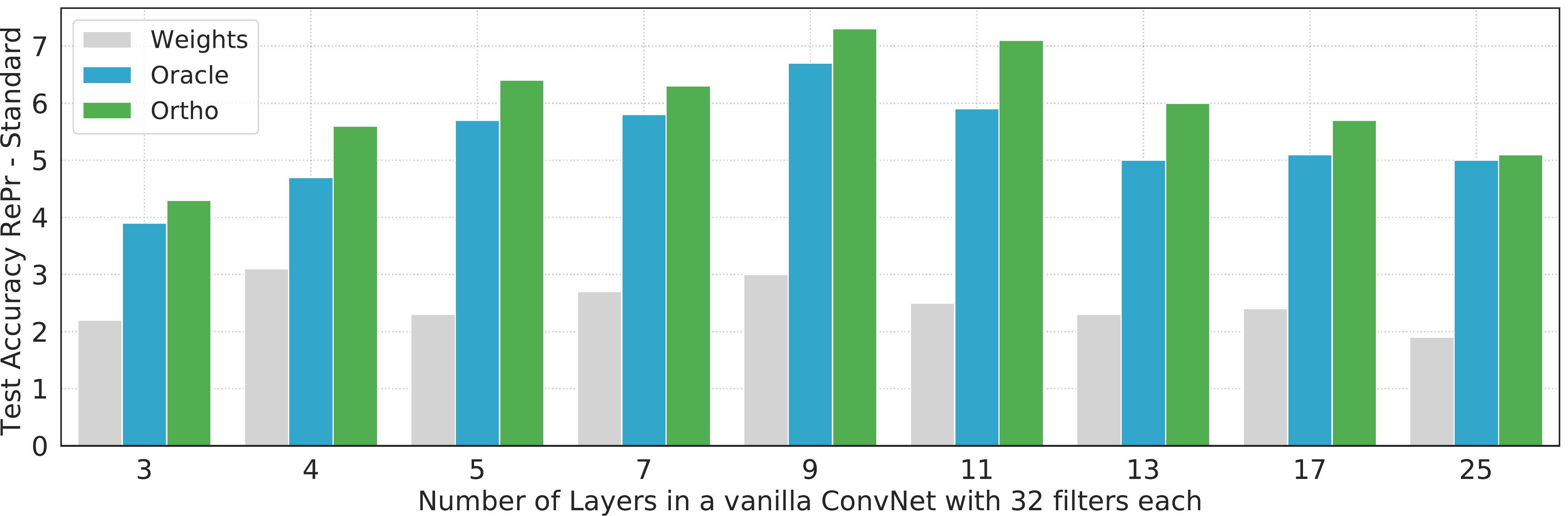}
   \caption{Accuracy improvement using RePr over standard training on Vanilla ConvNets across many layered networks [$C^n(32)$]}
   \label{fig:ablation_layers}
\end{figure}

\begin{table}[]
\center
\small
\begin{tabular}{cccccc}
\toprule
\multicolumn{6}{c}{\textbf{ResNet-20 on CIFAR-10}}                                                                                                                                                                                                                                                                                                     \\ \toprule
\multicolumn{2}{c}{Baseline}                                                                                 & \multicolumn{4}{c}{Various Training Schemes}                                                                                                                                                                                                    \\
\multicolumn{1}{c}{\textbf{\begin{tabular}[c]{@{}c@{}}Original\\ ~\cite{He2016DeepRL}\end{tabular}}}        & \multicolumn{1}{c|}{\textbf{\begin{tabular}[c]{@{}c@{}}Our \\ Impl\end{tabular}}} & \multicolumn{1}{c}{\textbf{\begin{tabular}[c]{@{}c@{}}DSD\\ ~\cite{Han2016DSDDT}\end{tabular}}}             & \multicolumn{1}{c|}{\textbf{\begin{tabular}[c]{@{}c@{}}BAN\\ ~\cite{Furlanello2018BornAN}\end{tabular}}} & \multicolumn{1}{c}{\textbf{\begin{tabular}[c]{@{}c@{}}RePr\\ Weights\end{tabular}}} & \multicolumn{1}{c}{\textbf{\begin{tabular}[c]{@{}c@{}}RePr \\ Ortho\end{tabular}}} \\ \hline
\multicolumn{1}{c}{8.7} & \multicolumn{1}{c|}{8.4}                                                         & \multicolumn{1}{c}{7.8} & \multicolumn{1}{c|}{8.2}          & 7.7                                                                                 & \textcolor{blue}{6.9}                     \\
\bottomrule
\end{tabular}
    \caption{Comparison of test error from using various techniques. }
   \label{tbl:resnet20}
\end{table}

We compare our training scheme with other similar schemes like BAN and DSD in table~\ref{tbl:resnet20}.
All three schemes were trained for three iterations \ie N=3. 
All models were trained for 150 epochs with similar learning rate schedule and initialization.
DSD and RePr (Weights) perform roughly the same function - sparsifying the model guided by magnitude, with the difference that DSD acts on individual weights, while RePr (Weights) acts on entire filters.
Thus, we observe similar performance between these techniques.
RePr (Ortho) outperforms the other techniques and is significantly cheaper to train compared to BAN, which requires N full training cycles.

Compared to modern architectures, vanilla ConvNets show significantly more inefficiency in the allocation of their feature representations. 
Thus, we find larger improvements from our method when applied to vanilla ConvNets, as compared to modern architectures. 
Table~\ref{tbl:cifar} shows test errors on CIFAR 10 \& 100. 
Vanilla CNNs with 32 filters each have high error compared to DenseNet or ResNet but their inference time is significantly faster. 
RePr training improves the relative accuracy of vanilla CNNs by $~8\%$ on CIFAR-10 and $~25\%$ on CIFAR-100.
The performance of baseline DenseNet and ResNet models is still better than vanilla CNNs trained with RePr, but these models incur more than twice the inference cost.
For comparison, we also consider a reduced DenseNet model with only $5$ layers, which has similar inference time to the 3-layer vanilla ConvNet. 
This model has many fewer parameters (by a factor of $11\times$) than the vanilla ConvNet, leading to significantly higher error rates, but we choose to equalize inference time rather than parameter count, due to the importance of inference time in many practical applications.
Figure~\ref{fig:ablation_layers} shows more results on vanilla CNNs with varying depth.
Vanilla CNNs start to overfit the data, as most filters converge to similar representation.
Our training scheme forces them to be different which reduces the overfitting (Figure~\ref{fig:ablation_percent} - right).
This is evident in the larger test error of 18-layer vanilla CNN with CIFAR-10 compared to 3-layer CNN.
With RePr training, $18$ layer model shows lower test error.

RePr is also able to improve the performance of ResNet and shallow DenseNet.
This improvement is larger on CIFAR-100, which is a $100$ class classification and thus is a harder task and requires more specialized filters.
Similarly, our training scheme shows bigger relative improvement on ImageNet, a $1000$ way classification problem.
Table~\ref{tbl:imagenet} presents top-1 test error on ImageNet~\cite{Russakovsky2015ImageNetLS} of various ConvNets trained using standard training and with RePr.
RePr was applied three times (N=3), and the table shows errors after each round.
We have attempted to replicate the results of the known models as closely as possible with suggested hyper-parameters and are within $\pm 1\%$ of the reported results.
More details of the training and hyper-parameters are provided in the supplementary material.
Each subsequent RePr leads to improved performance with significantly diminishing returns.
Improvement is more distinct in architectures which do not have skip connections, like Inception v1 and VGG and have lower baseline performance.
\begin{table}[]
\small
\begin{tabular}{ccccccc}
\toprule
                  &                                                                            & \multicolumn{1}{c}{}                                                                        & \multicolumn{2}{c}{\textbf{CIFAR-10}}                                                                & \multicolumn{2}{c}{\textbf{CIFAR-100}}     \\ \hline
Layers & \begin{tabular}[c]{@{}c@{}}Params\\ {\scriptsize($\times 000$)}\end{tabular} & \multicolumn{1}{c}{\begin{tabular}[c]{@{}c@{}}Inf. Time\\ {\scriptsize(relative)}\end{tabular}} & Std               & \multicolumn{1}{c}{RePr}                                       & Std & RePr               \\ \midrule
\multicolumn{7}{c}{Vanilla CNN {[}32 filters / layer{]}}                                                                                                                                                                                                                                                                                                   \\ \hline
\rowcolor[HTML]{EFEFEF} 
3                 & 20                                                                         & \multicolumn{1}{c}{\cellcolor[HTML]{EFEFEF}1.0}                                             & 27.9                       & \multicolumn{1}{c}{\cellcolor[HTML]{EFEFEF}{\color[HTML]{3166FF} 23.6}} & 52.8         & {\color[HTML]{3166FF} 41.8} \\
8                 & 66                                                                         & \multicolumn{1}{c}{1.7}                                                                     & 26.8                       & \multicolumn{1}{c}{{\color[HTML]{3166FF} 19.5}}                         & 50.9         & {\color[HTML]{3166FF} 36.8} \\
\rowcolor[HTML]{EFEFEF} 
13                & 113                                                                        & \multicolumn{1}{c}{\cellcolor[HTML]{EFEFEF}2.5}                                             & 26.6                       & \multicolumn{1}{c}{\cellcolor[HTML]{EFEFEF}{\color[HTML]{3166FF} 20.6}} & 51.0         & {\color[HTML]{3166FF} 37.9} \\
18                & 159                                                                        & \multicolumn{1}{c}{3.3}                                                                     & 28.2                       & \multicolumn{1}{c}{{\color[HTML]{3166FF} 22.5}}                         & 51.9         & {\color[HTML]{3166FF} 39.5} \\ \hline
\multicolumn{7}{c}{DenseNet {[}k=12{]}}                                                                                                                                                                                                                                                                                                            \\ \hline
\rowcolor[HTML]{EFEFEF} 
5                 & 1.7                                                                        & \multicolumn{1}{c}{\cellcolor[HTML]{EFEFEF}0.9}                                             & 39.4                       & \multicolumn{1}{c}{\cellcolor[HTML]{EFEFEF}{\color[HTML]{3166FF} 36.2}} & 43.5         & {\color[HTML]{3166FF} 40.9} \\
40                & 1016                                                                       & \multicolumn{1}{c}{8.0}                                                                     & 6.8                        & \multicolumn{1}{c}{{\color[HTML]{3166FF} 6.2}}                          & 26.4         & {\color[HTML]{3166FF} 25.2} \\
\rowcolor[HTML]{EFEFEF} 
100               & 6968                                                                       & \multicolumn{1}{c}{\cellcolor[HTML]{EFEFEF}43.9}                                            & {\color[HTML]{3166FF} 5.3} & \multicolumn{1}{c}{\cellcolor[HTML]{EFEFEF}{\color[HTML]{000000} 5.6}}  & 22.2         & {\color[HTML]{000000} 22.1} \\ \hline
\multicolumn{7}{c}{ResNet}                                                                                                                                                                                                                                                                                                                         \\ \hline
\rowcolor[HTML]{EFEFEF} 
20                & 269                                                                        & \multicolumn{1}{c}{\cellcolor[HTML]{EFEFEF}1.7}                                             & 8.4                        & \multicolumn{1}{c}{\cellcolor[HTML]{EFEFEF}{\color[HTML]{3166FF} 6.9}}  & 32.6         & {\color[HTML]{3166FF} 31.1} \\
32                & 464                                                                        & \multicolumn{1}{c}{2.2}                                                                     & 7.4                        & \multicolumn{1}{c}{{\color[HTML]{3166FF} 6.1}}                          & 31.4         & {\color[HTML]{3166FF} 30.1} \\
\rowcolor[HTML]{EFEFEF} 
110               & 1727                                                                       & \multicolumn{1}{c}{\cellcolor[HTML]{EFEFEF}7.1}                                             & 6.3                        & \multicolumn{1}{c}{\cellcolor[HTML]{EFEFEF}{\color[HTML]{3166FF} 5.4}}  & 27.5         & {\color[HTML]{3166FF} 26.4} \\
182               & 2894                                                                       & \multicolumn{1}{c}{11.7}                                                                    & 5.6                        & \multicolumn{1}{c}{{\color[HTML]{3166FF} 5.1}}                          & 26.0         & {\color[HTML]{3166FF} 25.3} \\ \bottomrule
\end{tabular}
\caption{Comparison of test error on Cifar-10 \& Cifar-100 of various ConvNets using Standard training vs RePr Training. Inf. Time shows the inference times for a single pass. All time measurements are relative to Vanilla CNN with three layers. Parameter count does not include the last fully-connected layer.}
   \label{tbl:cifar}
\end{table}

\begin{table}[]
\small
\begin{tabular}{cccccc}
\toprule
\multicolumn{6}{c}{\textbf{ImageNet}} \\
\midrule
      & Standard                             & \multicolumn{3}{c}{RePr Training}             &      Relative                       \\ \cline{3-5}
Model & Training & N=1 & N=2 & N=3 & Change \\ \midrule
\rowcolor[HTML]{EFEFEF} 
ResNet-18      & 30.41                       & 28.68        & 27.87        & {\color[HTML]{3166FF} 27.31} & -11.35                      \\
ResNet-34      & 27.50                       & 26.49        & 26.06        & {\color[HTML]{3166FF} 25.80} & -6.59                       \\
\rowcolor[HTML]{EFEFEF} 
ResNet-50      & 23.67                       & 22.79        & 22.51        & {\color[HTML]{3166FF} 22.37} & -5.81                       \\
ResNet-101     & 22.40                       & 21.70        & 21.51        & {\color[HTML]{3166FF} 21.40} & -4.67                       \\
\rowcolor[HTML]{EFEFEF} 
ResNet-152     & 21.51                       & 20.99        & 20.79        & {\color[HTML]{3166FF} 20.71} & -3.86                       \\
VGG-16         & 31.30                       & 27.76        & 26.45        & {\color[HTML]{3166FF} 25.50} & -22.75                      \\
\rowcolor[HTML]{EFEFEF} 
Inception v1   & 31.11                       & 29.41        & 28.47        & {\color[HTML]{3166FF} 28.01} & -11.07                      \\
Inception v2   & 27.60                       & 27.15        & 26.95        & {\color[HTML]{3166FF} 26.80} & -2.99                       \\ \bottomrule
\end{tabular}
   \caption{Comparison of test error (Top-1) on ImageNet with different models at various stages of RePr. N=1, N=2, N=3 are results after each round of RePr.}
   \label{tbl:imagenet}
\end{table}

Our model improves upon other computer vision tasks that use similar ConvNets. We present a small sample of results from visual question answering and object detection tasks.
Both these tasks involve using ConvNets to extract features, and RePr improves their baseline results.

\textbf{Visual Question Answering}
In the domain of visual question answering (VQA), a model is provided with an image and question (as text) about that image, and must produce an answer to that question.
Most of the models that solve this problem use standard ConvNets to extract image features and an LSTM network to extract text features.
These features are then fed to a third model which learns to select the correct answer as a classification problem. State-of-the-art models use an attention layer and intricate mapping between features. We experimented with a more standard model where image features and language features are fed to a Multi-layer Perceptron with a softmax layer at the end that does $1000$-way classification over candidate answers.
Table~\ref{tbl:vqa} provides accuracy on VQAv1 using VQA-LSTM-CNN model~\cite{Antol2015VQAVQ}.
Results are reported for Open-Ended questions, which is a harder task compared to multiple-choice questions.
We extract image features from Inception-v1, trained using standard training and with RePr (Ortho) training, and then feed these image features and the language embeddings (GloVe vectors) from the question, to a two layer fully connected network. 
Thus, the only difference between the two reported results~\ref{tbl:vqa} is the training methodology of Inception-v1.

\begin{table}[]
\center
\begin{tabular}{ccccc}
        \toprule
         & All & Yes/No & Other & Number \\ \midrule
Standard & 60.3    &    81.4    &  47.6     &  37.2      \\
RePr (Ortho)    & {\color[HTML]{3166FF} 64.6}    &  {\color[HTML]{3166FF} 83.4}      &  {\color[HTML]{3166FF} 54.5}     &  37.2     \\ \bottomrule
\end{tabular}
\caption{Comparison of Standard Training and RePr on VQA using VQA-LSTM-CNN model}
\label{tbl:vqa}
\end{table}

\textbf{Object Detection}
For object detection, we experimented with Faster R-CNN using ResNet $50$ and $101$ pretrained on ImageNet.
We experimented with both Feature Pyramid Network and baseline RPN with \textit{c4} conv layer.
We use the model structure from Tensorpack~\cite{wu2016tensorpack}, which is able to reproduce the reported mAP scores.
The model was trained on 'trainval35k + minival' split of COCO dataset (2014). 
Mean Average Precision (mAP) is calculated at ten IoU thresholds from $0.5$ to $0.95$. 
mAP for the boxes obtained with standard training and RePr training is shown in the table~\ref{tbl:rpn}.

\begin{table}[H]
\center
\begin{tabular}{ccccc}
        \toprule
         & \multicolumn{2}{l}{ResNet-50} & \multicolumn{2}{l}{ResNet-101} \\
         & RPN           & FPN           & RPN           & FPN           \\ \midrule
Standard  &      38.1  & 38.2      &  40.7     & 41.7\\
RePr (Ortho) &   {\color[HTML]{3166FF} 41.1}  & {\color[HTML]{3166FF} 42.3}     &   {\color[HTML]{3166FF} 43.5}     & {\color[HTML]{3166FF} 44.5}       \\ \bottomrule
\end{tabular}
\caption{mAP scores with Standard and RePr (Ortho) training for object detection with ResNet the ConvNet (RPN on C4)}
\label{tbl:rpn}
\end{table}

\section{Conclusion} \label{sec:conclusion}
We have introduced RePr, a training paradigm which cyclically drops and relearns some percentage of the least expressive filters.
After dropping these filters, the pruned sub-model is able to recapture the lost features using the remaining parameters, allowing a more robust and efficient allocation of model capacity once the filters are reintroduced.
We show that a reduced model needs training before re-introducing the filters, and careful selection of this training duration leads to substantial gains. 
We also demonstrate that this process can be repeated with diminishing returns.

Motivated by prior research which highlights inefficiencies in the feature representations learned by convolutional neural networks, we further introduce a novel \textit{inter}-filter orthogonality metric for ranking filter importance for the purpose of RePr training, and demonstrate that this metric outperforms established ranking metrics.
Our training method is able to significantly improve performance in under-parameterized networks by ensuring the efficient use of limited capacity, and the performance gains are complementary to knowledge distillation.
Even in the case of complex, over-parameterized network architectures, our method is able to improve performance across a variety of tasks.

\section{Acknowledgement}
First author would like to thank NVIDIA and Google for donating hardware resources partially used for this research. 
He would also like to thank Nick Moran, Solomon Garber and Ryan Marcus for helpful comments.

{\small
\bibliographystyle{unsrt}
\bibliographystyle{ieee}
\bibliography{egbib}
}

\section*{Appendix}

\subsection*{ImageNet Training Details}
Training large models such as ResNet, VGG or Inception (as discussed in Table 4) can be difficult and models may not always converge to similar optima across training runs. 
With our RePr training scheme, we observed that large values of $p_\%$ can sometimes produce collapse upon re-introduction of dropped filters.
On analysis, we found that this was due to large random activations from the newly initialized filters. 
This can be overcome by initializing the new filters with relatively small values. 

Another trick that minimizes this problem is to also re-initialize the corresponding kernels of next layer for a given filter.
Consider a filter $f$ at layer $\ell$. 
The activations from this filter $f$ become input to a kernel of every filter of the next layer $\ell + 1$.
If the filter $f$ is pruned, and then re-initialized, then all those kernels in layer $\ell + 1$ should also be initialized to small random values, as the features they had learned to process no longer exist.
This prevents new activations these kernels (which are currently random) from dominating the activations from other kernels.

Pruning significant number of filters at one iteration could lead to instability in training. 
This is mostly due to changes in running mean/variance of BatchNorm parameters.
To overcome this issue, filters can be pruned over multiple mini-batches. 
There is no need to re-evaluate the rank, as it does not change significantly with few iterations.
Instability of training is compounded in DenseNet, due to the dense connections.
Removing multiple filters leads to significant changes to the forward going dense connections, and they impact all the existing activations.
One way to overcome this is to decay the filter weights over multiple iterations to a very small norm before removing the filter all together from the network.
Similarly Squeeze-and-Excitation Networks\footnote{Hu, J., Shen, L., \& Sun, G. (2017). Squeeze-and-Excitation Networks. CoRR, abs/1709.01507.} are also difficult to prune, because they maintain learned scaling parameters for activations from all the filters.
Unlike, BatchNorm, it is not trivial to remove the corresponding scaling parameters, as they are part of a fully connected layer. Removing this value would change the network structure and also relative scaling of all the other activations.

It is also possible to apply RePr to a pre-trained model. This is especially useful for ImageNet, where the cost of training from scratch is high.
Applying RePr to a pre-trained model is able to produce some improvement, but is not as effective as applying RePr throughout training.
Careful selection of the fine-tuning learning rate is necessary to minimize the required training time.
Our experiments show that using adaptive LR optimizers such as Adam might be more suited for fine-tuning from pre-trained weights.

\subsection*{Hyper-parameters}
All ImageNet models were trained using Tensorflow with Tesla V100 and model definitions were obtained from the official TF repository\footnote{tensorflow/contrib/slim/python/slim/nets}. 
Images were augmented with brightness (0.6 to 1.4), contrast (0.6 to 1.4), saturation (0.4), lightning (0.1), random center crop and horizontal flip. During the test, images were tested on center $224 \times 224$ crop. Most models were trained with a batch size of 256, but the large ones like ResNet-101, ResNet-152 and Inception-v2 were trained with a batch size of 128. 
Depending upon the implementation, RePr may add its own non-trainable variables, which will take up GPU memory, thus requiring the use smaller batch size than that originally reported by other papers. 
Models with batch sizes of 256 were trained using SGD with a learning rate of 0.1 for the first 30 epochs, 0.01 for the next 30 epochs, and 0.001 for the remaining epochs.
For models with batch sizes of 128, these learning rates were correspondingly reduced by half.
For ResNet models convolutional layers were initialized with MSRA initialization with FAN OUT (scaling=$2.0$), and fully connected layer was initialized with Random Normal (standard deviation =$0.01$).
\begin{figure*}
\center
   \includegraphics[width=1.0\linewidth]{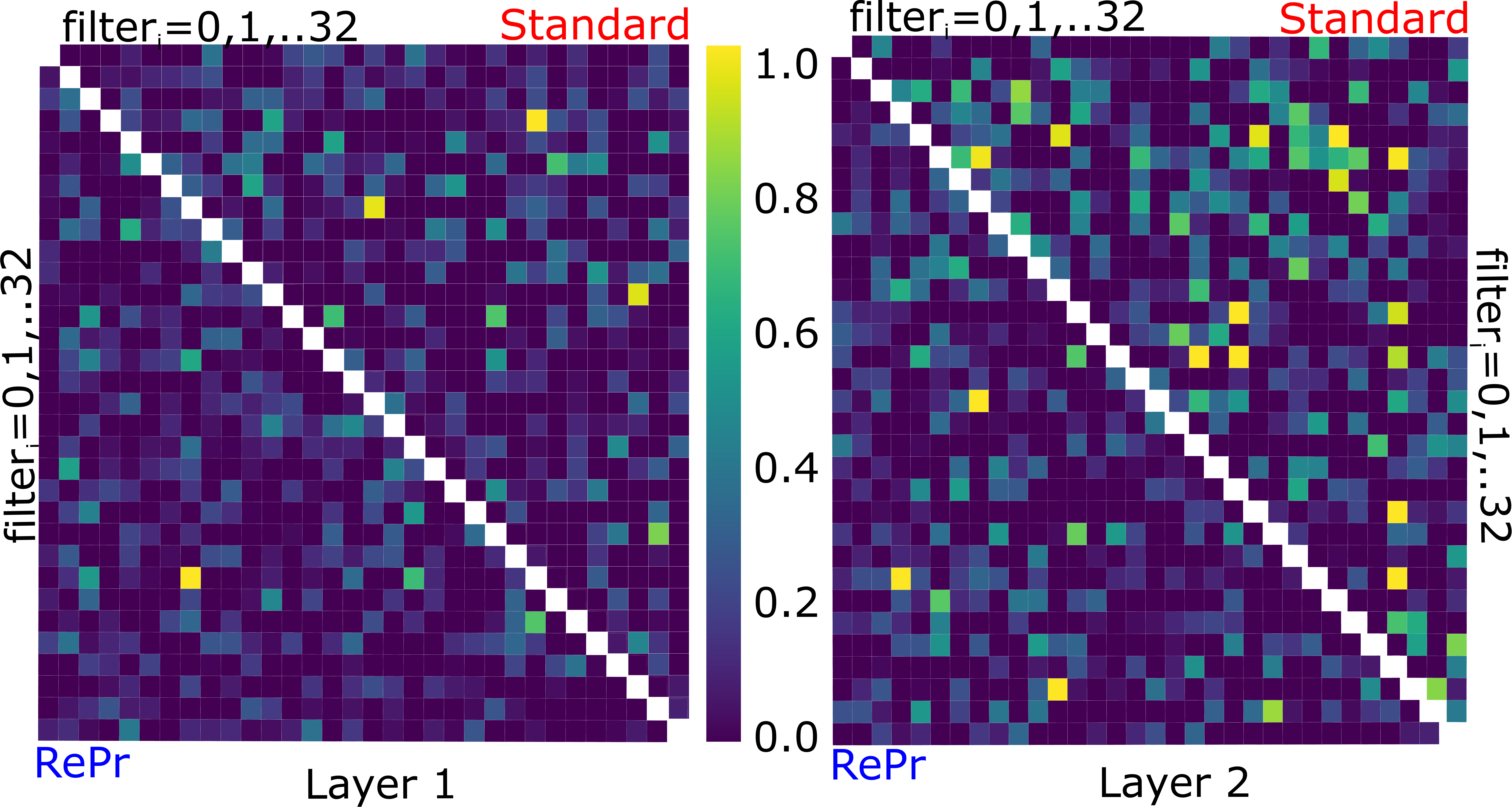}
 
   \caption{Comparison of filter correlations with RePr and Standard Training.}
 
\end{figure*}
\end{document}